\documentclass{ieeeaccess}
\usepackage{cite}
\usepackage{amsmath,amssymb,amsfonts}
\usepackage{algpseudocode}
\usepackage{graphicx}
\usepackage{textcomp}
\def\BibTeX{{\rm B\kern-.05em{\sc i\kern-.025em b}\kern-.08em
    T\kern-.1667em\lower.7ex\hbox{E}\kern-.125emX}}

\usepackage{amstext}
\usepackage{cancel}
\usepackage{multirow}
\usepackage{rotating}
\usepackage{placeins}
\usepackage{array}

\usepackage{algorithm}

\usepackage[T1]{fontenc}
\usepackage{pslatex}
\usepackage{contour}
\usepackage{hyperref}

\usepackage{subfig}
\usepackage{array} 

\usepackage{color,soul}
\soulregister\cite7
\soulregister\ref7
\soulregister\pageref7

\usepackage[table]{xcolor}

\usepackage{float}
\usepackage{afterpage}

\newcommand\imgsize{55mm}
\newcommand\vspacetable{-10pt}

\newcommand{\aggresults}{
    
    \begin{table*}[p]
        \centering
        \caption{HiER compared to the state-of-the-art across all tasks. For the reward, there is no universal desirable target, thus there is no OG value. The column-wise best results are marked in bold. Both HiER version outperform their corresponding baseline. HiER~[HER] yields the best performance in all metrics.}
        \label{tab:results_agg_alltasks}
        \begin{tabular}{c|c|cccc|ccc}
                & & \multicolumn{4}{c}{Success rate} & \multicolumn{3}{c}{Reward} \\
                
              & \rotatebox{90}{HER} \rotatebox{90}{HiER} & Mean $\uparrow$ & Median $\uparrow$ & IQM $\uparrow$ & OG $\downarrow$ & Mean $\uparrow$ & Median $\uparrow$ & IQM $\uparrow$ \\
               \hline
                \multirow{2}{*}{Baselines}  &  - -  &  0.19 & 0.10 & 0.09 & 0.81 & -111.56 &  -48.2 & -48.91 \\
             &  \checkmark - & 0.57 & 0.50 & 0.56 & 0.43 & -87.50 & -43.19 & -43.70 \\
             \hline
             \multirow{2}{*}{HiER} & - \checkmark &  0.44 & 0.38 & 0.38 & 0.56 & -98.72 & -40.96 & -42.28 \\
             &  \checkmark \checkmark  & \textbf{0.75} & \textbf{0.80} & \textbf{0.83} & \textbf{0.25} & \textbf{-73.14} & \textbf{-31.35} & \textbf{-32.48} \\
        \end{tabular}
    \end{table*}
    
    \begin{figure*}[p]
    \centering
    \includegraphics[width=1.0\textwidth]{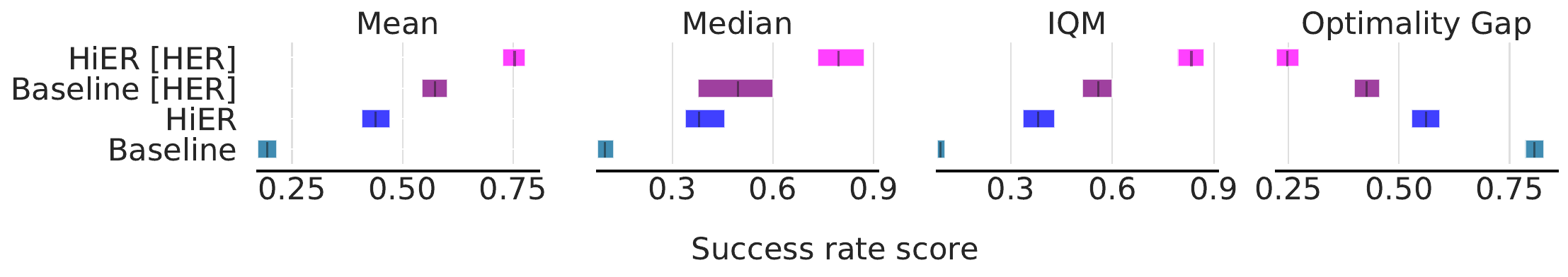}
    \caption{HiER compared to the state-of-the-art across all tasks with 95\% CIs. Both HiER version outperform their corresponding baseline. HiER~[HER] yields the best performance in all metrics. The point estimates are presented in Tab.~\ref{tab:results_agg_alltasks}}
    \label{fig:all_agg_metrics}
    \end{figure*}
    
    \begin{figure*}[p]
        \centering
    
        \subfloat{%
          \includegraphics[width=0.4\linewidth]{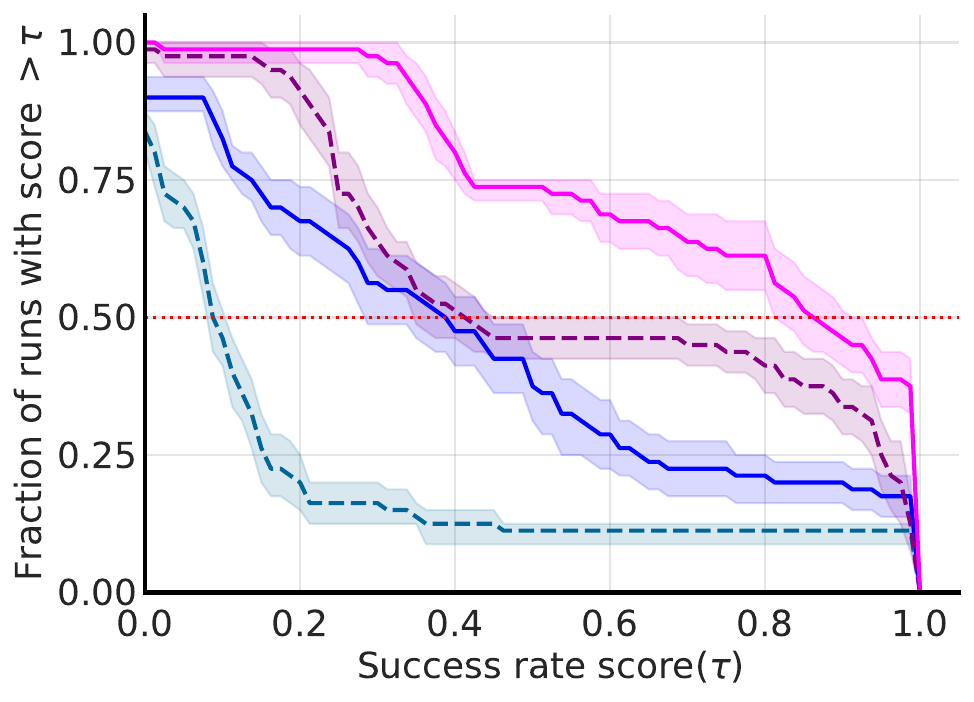}
        }
        \subfloat{%
          \includegraphics[width=0.4\linewidth]{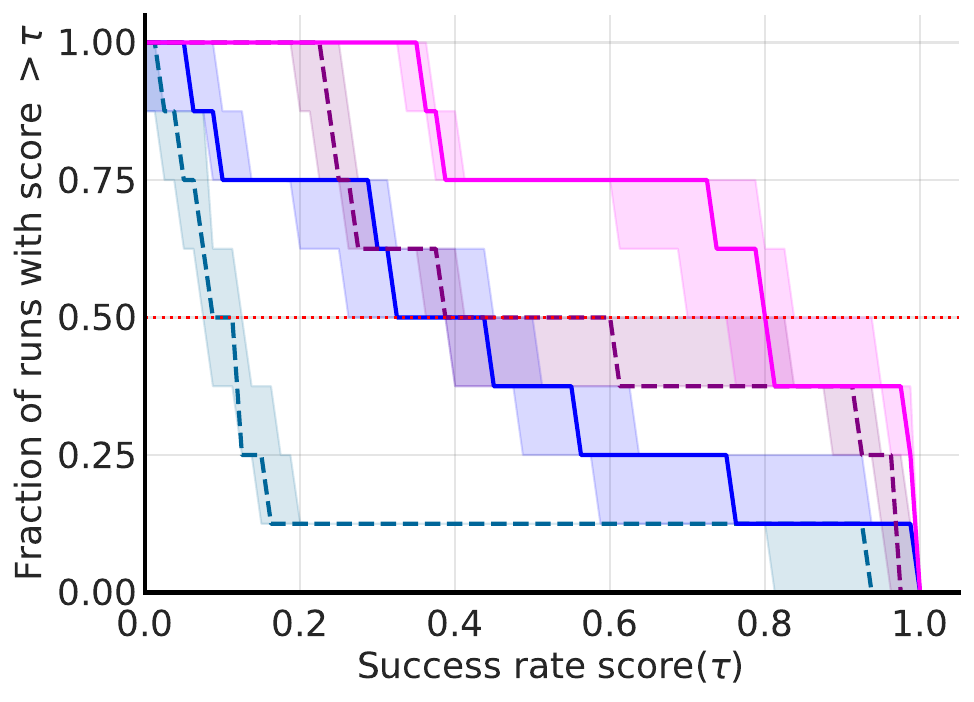}
        }
        \hfill
    
        \subfloat{%
          \includegraphics[width=\textwidth]{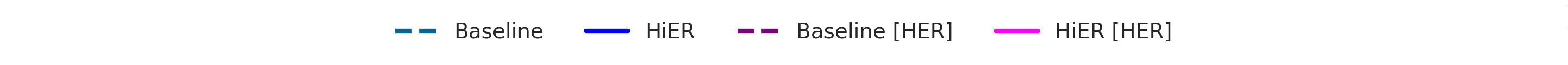}
        }
    
        \caption{Performance profiles across all tasks with 95\% CIs. \textbf{Left}: run-score distribution, \textbf{right}: average-score distribution. The red-dotted line shows the median values while the areas under the performance profiles correspond to the mean values (comparing with Tab.~\ref{tab:results_agg_alltasks}, the average-score distribution needs to be examined). Both HiER and HiER~[HER] have stochastic dominance over their corresponding baselines.}
        \label{fig:all_performance_profile}
    \end{figure*}
    
    \begin{figure*}[p]
        \centering
        \includegraphics[angle=0,width=0.4\linewidth]{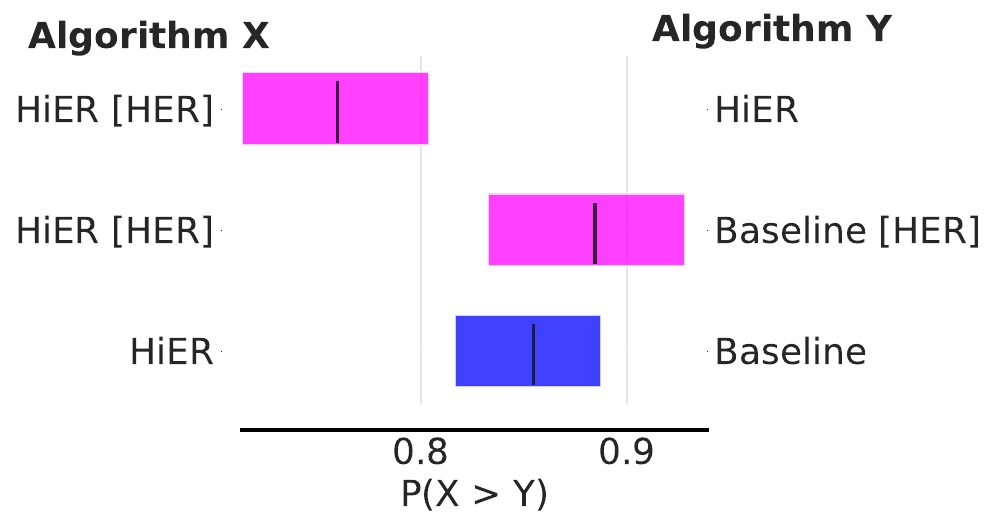}
        \caption{Probability of improvement of HiER versions compared to their corresponding baselines and themselves across all tasks with 95\% CIs. The average probabilities from top to bottom are the following: 0.76, 0.88, and 0.85.}
        \label{fig:all_probability_hier}
    \end{figure*}
    
}

\newcommand{\pandafigs}{

    \begin{figure*}[p]
        \centering
        \subfloat{%
          \includegraphics[width=\imgsize]{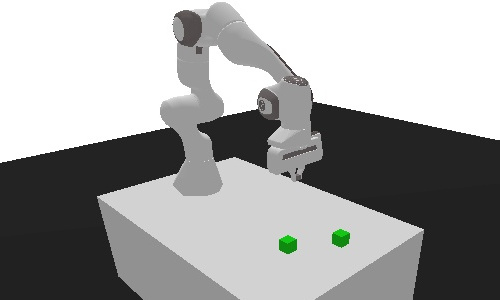}
        }
        \hfill
        \subfloat{%
          \includegraphics[width=\imgsize]{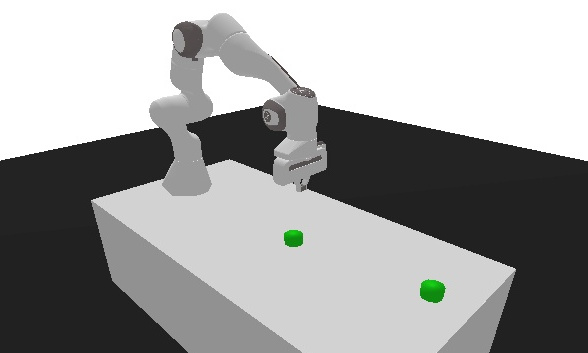}
        }
        \hfill
        \subfloat{%
          \includegraphics[width=\imgsize]{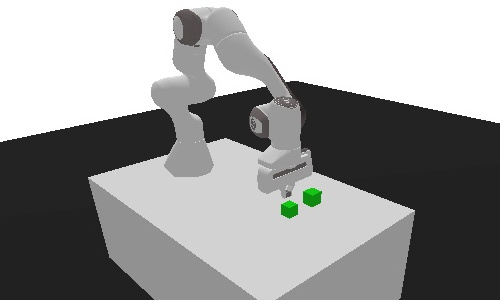}
        }
        
        \subfloat{%
          \includegraphics[width=\imgsize]{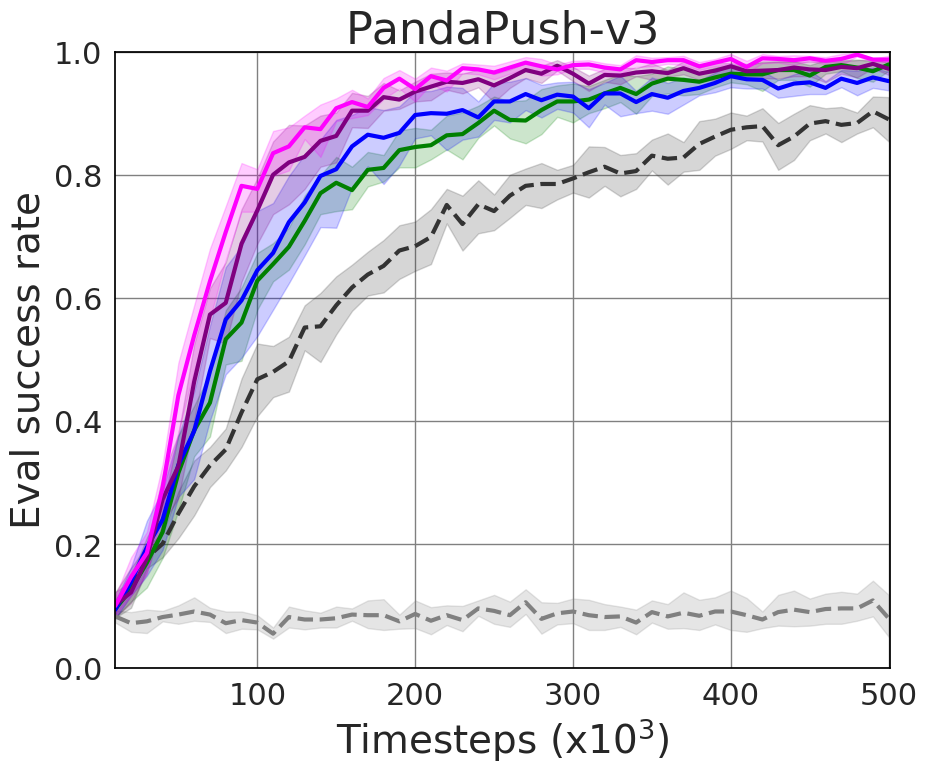}
        }
        \hfill
        \subfloat{%
          \includegraphics[width=\imgsize]{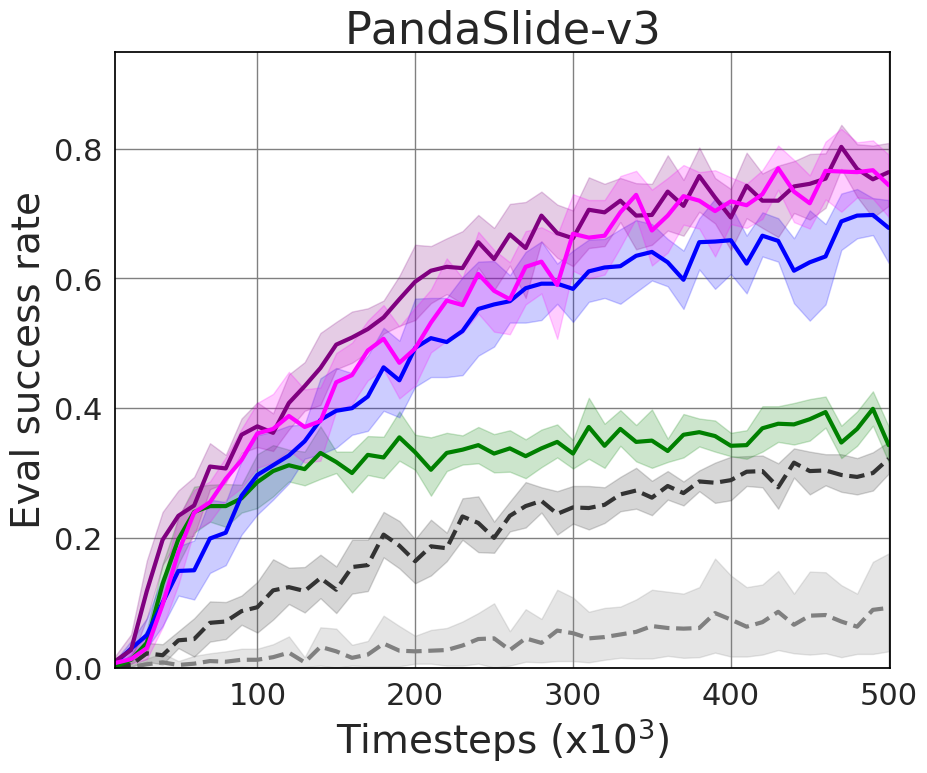}
        }
        \hfill
        \subfloat{%
          \includegraphics[width=\imgsize]{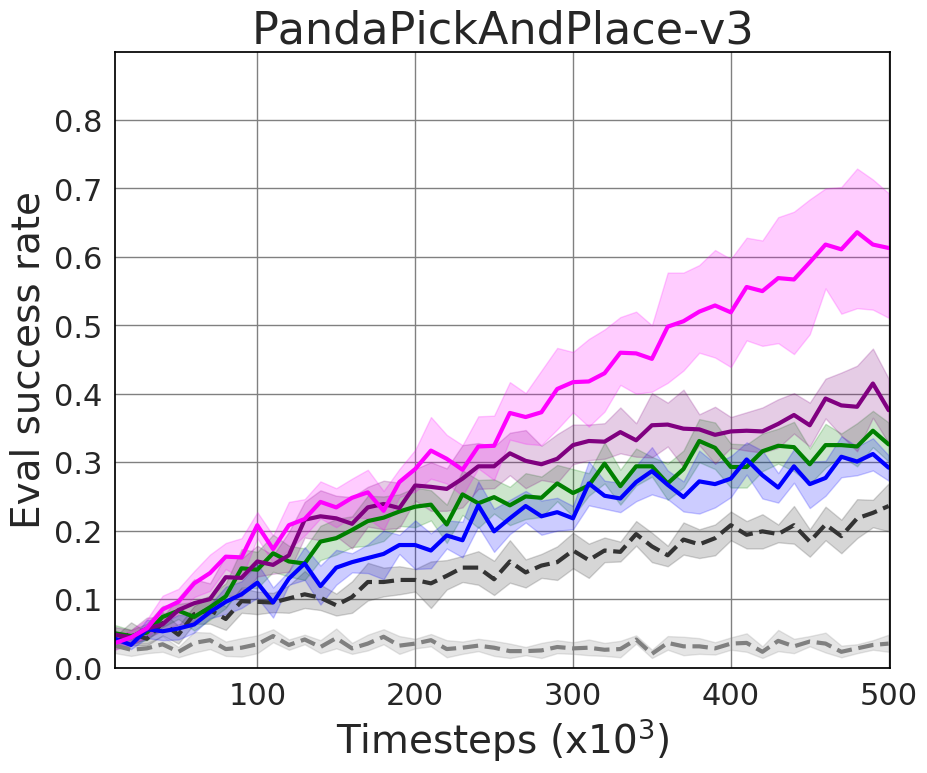}
        }
    
        \subfloat{%
          \includegraphics[width=\textwidth]{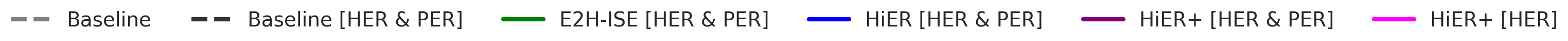}
        }
    
        \caption{Learning curves of HiER and HiER+ with E2H-ISE compared to the state-of-the-art based on success rates on the push, slide, and pick-and-place tasks of the Panda-Gym robotic benchmark with 95\% CIs.}
        \label{fig:result_panda_tasks}
    \end{figure*}

    \begin{figure*}[p]
    \centering
    \includegraphics[width=1.0\textwidth]{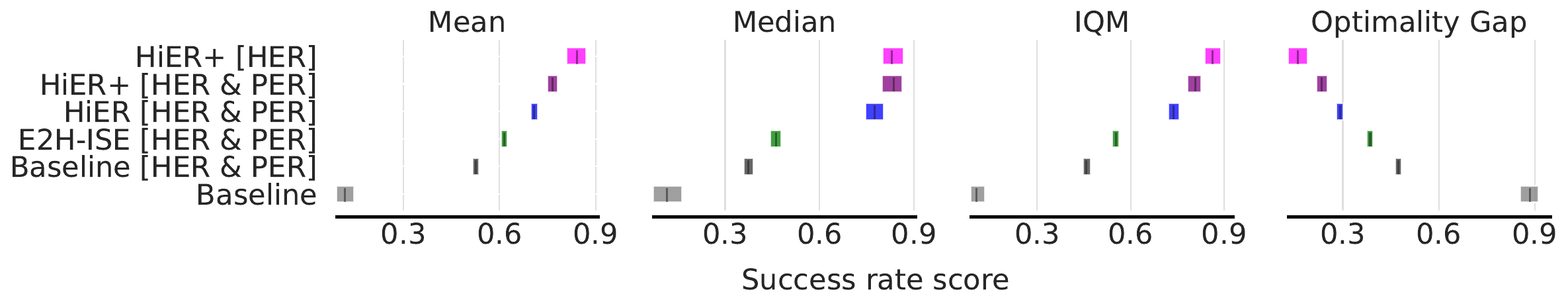}
    \caption{Aggregate metrics on the push, slide, and pick-and-place tasks of the Panda-Gym robotic benchmark with 95\% CIs. HiER (blue) and both versions of HiER+ (purple and magenta) significantly outperform the baselines (gray). E2H-ISE alone could slightly improve the performance of the baseline.}
    \label{fig:panda_agg_metrics}
    \end{figure*}
    
    \begin{figure*}[p]
        \centering
    
        \subfloat{%
          \includegraphics[width=0.35\linewidth]{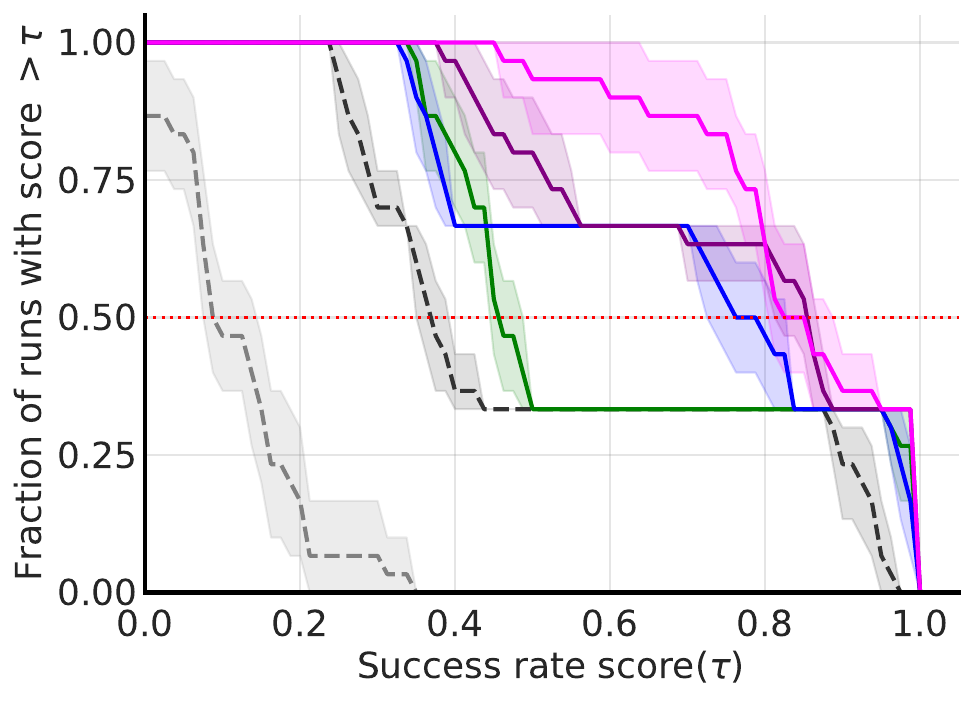}
        }
        \subfloat{%
          \includegraphics[width=0.55\linewidth]{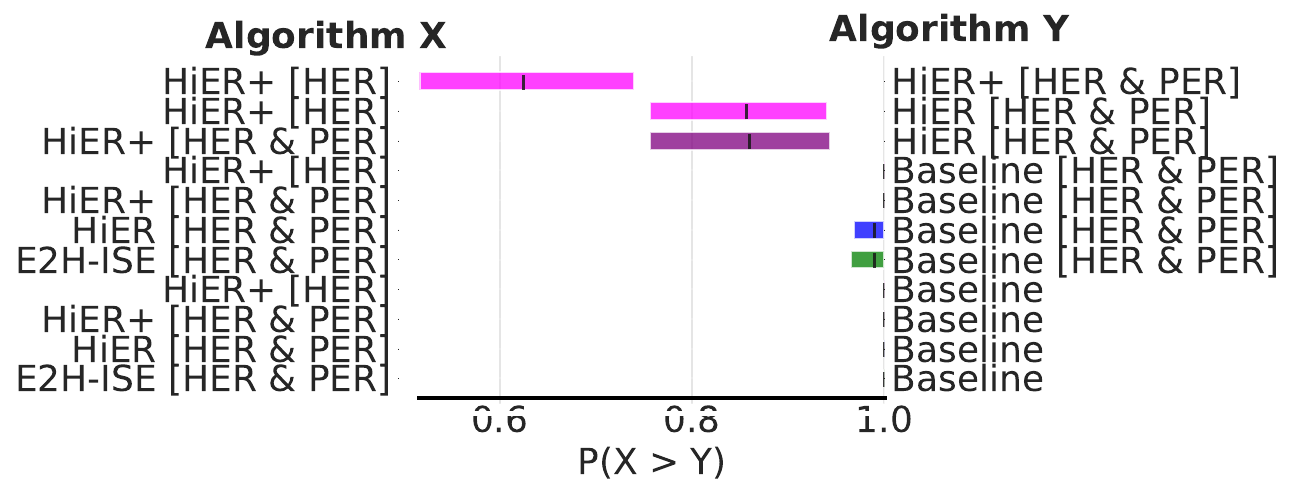}
        }
        \hfill
    
        \subfloat{%
          \includegraphics[width=\textwidth]{figures/GrandTest/Z_1212_A_GT_01_legend.png}
        }
        \caption{\textbf{Left:} performance profiles (run-score distribution) on the push, slide, and pick-and-place tasks of the Panda-Gym robotic benchmark with 95\% CIs. \textbf{Right:} Probability of improvement on the push, slide, and pick-and-place tasks of the Panda-Gym robotic benchmark with 95\% CIs. The average probabilities from top to bottom: 0.625, 0.857, 0.86, 1.0, 1.0, 0.99, 0.99, 1.0, 1.0, and 1.0.}
        \label{fig:panda_performance_profile_and_prob}
    \end{figure*}

}
\newcommand{\pandatabs}{

    \begin{table*}[p]
        \centering
         \caption{Simplified summary of our results on the push, slide, and pick-and-place tasks of the Panda-Gym robotic benchmark based on success rates. The column-wise best results are marked in bold. The full table with all the configurations is presented in Tab.~\ref{tab:results_sota}.}
      \label{tab:results_sota_summary}
        \begin{tabular}{c|cccccc}
              & \multicolumn{6}{c}{\texttt{PandaPush-v3} | \texttt{PandaSlide-v3} | \texttt{PandaPickAndPlace-v3} } \\
              &  Mean $\uparrow$ & Median $\uparrow$ & IQM $\uparrow$ & OG $\downarrow$ & Max $\uparrow$ & Std $\downarrow$ \\
       \hline
          Baseline~[HER] & 0.97 | 0.38 | 0.27 & 0.98 | 0.37 | 0.28 & 0.97 | 0.37 | 0.27 & 0.03 | 0.62 | 0.73 & 0.99 | 0.45 | 0.32 & 0.02 | \textbf{0.04} | 0.03 \\
     HiER~[HER] & \textbf{1.00} | 0.79 | 0.39 & \textbf{1.00} | \textbf{0.81} | 0.39 & \textbf{1.00} | 0.81 | 0.39 & \textbf{0.00} | 0.21 | 0.61 & \textbf{1.00} | 0.91 | 0.42 & \textbf{0.00} | 0.09 | \textbf{0.02} \\
     HiER+~[HER] & \textbf{1.00} | \textbf{0.83} | \textbf{0.69} & \textbf{1.00} | \textbf{0.81} | \textbf{0.74} & \textbf{1.00} | \textbf{0.82} | \textbf{0.71} & \textbf{0.00} | \textbf{0.17} | \textbf{0.31} & \textbf{1.00} | \textbf{0.95} | \textbf{0.90} & \textbf{0.00} | 0.05 | 0.14 \\
        \end{tabular}
        \vspace{\vspacetable} % Adjust space after table
    \end{table*}

    \begin{table*}[p]
        \centering
        \caption{HiER and HiER+ compared to the state-of-the-art based on success rates on the Panda-Gym robotic benchmark. On the left side of the header, the components of the specific algorithm are displayed (HER, PER, ISE, HiER). The column-wise best results are marked in bold.}
        \label{tab:results_sota}
        \begin{tabular}{c|c|cccccc}
              & & \multicolumn{6}{c}{\texttt{PandaPush-v3} | \texttt{PandaSlide-v3} | \texttt{PandaPickAndPlace-v3} } \\
              & \rotatebox{90}{HER} \rotatebox{90}{PER} \rotatebox{90}{ISE} \rotatebox{90}{HiER} & Mean $\uparrow$ & Median $\uparrow$ & IQM $\uparrow$ & OG $\downarrow$ & Max $\uparrow$ & Std $\downarrow$ \\
       \hline
        \multirow{4}{*}{\rotatebox{90}{Baselines} }  &  -  -  - -  & 0.16 | 0.12 | 0.07 & 0.15 | 0.05 | 0.07 & 0.15 | 0.08 | 0.07 & 0.84 | 0.88 | 0.93 & 0.21 | 0.34 | 0.09 & 0.03 | 0.13 | \textbf{0.01} \\
     &  \checkmark  -  - -  & 0.97 | 0.38 | 0.27 & 0.98 | 0.37 | 0.28 & 0.97 | 0.37 | 0.27 & 0.03 | 0.62 | 0.73 & 0.99 | 0.45 | 0.32 & 0.02 | 0.04 | 0.03 \\
     &  -  \checkmark  - -  & 0.26 | 0.25 | 0.08 & 0.25 | 0.27 | 0.09 & 0.25 | 0.27 | 0.08 & 0.74 | 0.75 | 0.92 & 0.43 | 0.42 | 0.10 & 0.07 | 0.14 | \textbf{0.01} \\
     &  \checkmark  \checkmark  - -  & 0.93 | 0.37 | 0.28 & 0.94 | 0.37 | 0.28 & 0.93 | 0.37 | 0.27 & 0.07 | 0.63 | 0.72 & 0.97 | 0.43 | 0.33 & 0.03 | \textbf{0.02} | 0.02 \\
       \hline
       \hline
        \multirow{4}{*}{\rotatebox{90}{HiER} }  &  -  -  - \checkmark  & 0.44 | 0.29 | 0.09 & 0.44 | 0.28 | 0.09 & 0.44 | 0.29 | 0.09 & 0.56 | 0.71 | 0.91 & 0.57 | 0.39 | 0.11 & 0.09 | 0.07 | \textbf{0.01} \\
     &  \checkmark  -  - \checkmark  & \textbf{1.00} | 0.79 | 0.39 & \textbf{1.00} | 0.81 | 0.39 & \textbf{1.00} | 0.81 | 0.39 & \textbf{0.00} | 0.21 | 0.61 & \textbf{1.00} | 0.91 | 0.42 & \textbf{0.00} | 0.09 | 0.02 \\
     &  -  \checkmark  - \checkmark  & 0.80 | 0.41 | 0.13 & 0.88 | 0.42 | 0.13 & 0.83 | 0.44 | 0.13 & 0.20 | 0.59 | 0.87 & 0.98 | 0.66 | 0.16 & 0.15 | 0.17 | 0.02 \\
     &  \checkmark  \checkmark  - \checkmark  & 0.98 | 0.78 | 0.37 & 0.99 | 0.78 | 0.37 & 0.99 | 0.78 | 0.37 & 0.02 | 0.22 | 0.63 & \textbf{1.00} | 0.83 | 0.39 & 0.01 | 0.05 | 0.02 \\
       \hline
       \hline
        \multirow{4}{*}{\rotatebox{90}{ISE} }  &  -  -  \checkmark  -  & 0.85 | 0.45 | 0.25 & 0.85 | 0.45 | 0.25 & 0.86 | 0.45 | 0.25 & 0.15 | 0.55 | 0.75 & 0.95 | 0.47 | 0.30 & 0.06 | \textbf{0.02} | 0.03 \\
     &  \checkmark  -  \checkmark  -  & \textbf{1.00} | 0.45 | 0.42 & \textbf{1.00} | 0.45 | 0.43 & \textbf{1.00} | 0.45 | 0.43 & \textbf{0.00} | 0.55 | 0.58 & \textbf{1.00} | 0.52 | 0.53 & 0.01 | 0.03 | 0.04 \\
     &  -  \checkmark  \checkmark  -  & 0.83 | 0.44 | 0.31 & 0.83 | 0.44 | 0.30 & 0.83 | 0.44 | 0.30 & 0.17 | 0.56 | 0.69 & 0.89 | 0.52 | 0.36 & 0.04 | 0.04 | 0.03 \\
     &  \checkmark  \checkmark  \checkmark  -  & 0.99 | 0.46 | 0.39 & \textbf{1.00} | 0.46 | 0.38 & \textbf{1.00} | 0.46 | 0.39 & 0.01 | 0.54 | 0.61 & \textbf{1.00} | 0.50 | 0.44 & 0.01 | 0.03 | 0.03 \\
       \hline
        \multirow{4}{*}{\rotatebox{90}{HiER+} }  &  -  -  \checkmark  \checkmark  & 0.98 | 0.53 | 0.33 & 0.99 | 0.48 | 0.32 & 0.99 | 0.49 | 0.32 & 0.02 | 0.47 | 0.67 & \textbf{1.00} | 0.76 | 0.39 & 0.01 | 0.12 | 0.03 \\
     &  \checkmark  -  \checkmark  \checkmark  & \textbf{1.00} | 0.83 | \textbf{0.69} & \textbf{1.00} | 0.81 | \textbf{0.74} & \textbf{1.00} | 0.82 | \textbf{0.71} & \textbf{0.00} | 0.17 | \textbf{0.31} & \textbf{1.00} | \textbf{0.95} | \textbf{0.90} & \textbf{0.00} | 0.05 | 0.14 \\
     &  -  \checkmark  \checkmark  \checkmark  & 0.98 | 0.51 | 0.41 & 0.98 | 0.49 | 0.40 & 0.98 | 0.50 | 0.40 & 0.02 | 0.49 | 0.59 & \textbf{1.00} | 0.65 | 0.50 & 0.02 | 0.07 | 0.05 \\
     &  \checkmark  \checkmark  \checkmark  \checkmark  & \textbf{1.00} | \textbf{0.84} | 0.47 & \textbf{1.00} | \textbf{0.86} | 0.45 & \textbf{1.00} | \textbf{0.85} | 0.46 & \textbf{0.00} | \textbf{0.16} | 0.53 & \textbf{1.00} | 0.88 | 0.55 & \textbf{0.00} | 0.05 | 0.06 \\
        \end{tabular}
        \vspace{\vspacetable} % Adjust space after table
    \end{table*}

    \begin{table*}[p]
        \centering
        \caption{HiER and HiER+ compared to the state-of-the-art based on the evaluation rewards on the Panda-Gym robotic benchmark. On the left side of the header, the components of the specific algorithm are displayed (HER, PER, ISE, HiER). The desired performance scores for the OG metric are -10, -20, and -30 for the push, slide, and pick-and-place tasks respectively. The column-wise best results are marked in bold.}
        \label{tab:results_sota_reward}
        \begin{tabular}{c|c|cccccc}
              & & \multicolumn{6}{c}{\texttt{PandaPush-v3} | \texttt{PandaSlide-v3} | \texttt{PandaPickAndPlace-v3} } \\
              & \rotatebox{90}{HER} \rotatebox{90}{PER} \rotatebox{90}{ISE} \rotatebox{90}{HiER} & Mean $\uparrow$ & Median $\uparrow$ & IQM $\uparrow$ & OG $\downarrow$ & Max $\uparrow$ & Std $\downarrow$ \\
       \hline
        \multirow{4}{*}{\rotatebox{90}{Baselines} }  &  -  -  - -  & -46.2 | -46.7 | -48.2 & -46.2 | -49.0 | -48.5 & -46.4 | -48.0 | -48.3 & 36.2 | 26.7 | 18.2 & -41.0 | -37.6 | -46.5 & \hspace{1.25pt} 2.4 | \hspace{1.25pt} 4.3 | \hspace{1.25pt} 1.0 \\
     &  \checkmark  -  - -  & -11.4 | -39.2 | -41.8 & -11.2 | -38.5 | -41.1 & -11.2 | -38.7 | -41.5 & \hspace{1.25pt} 1.5 | 19.2 | 11.8 & \hspace{1.25pt} -9.6 | -36.5 | -39.5 & \hspace{1.25pt} 1.3 | \hspace{1.25pt} 2.0 | \hspace{1.25pt} 1.9 \\
     &  -  \checkmark  - -  & -40.8 | -43.5 | -48.8 & -41.6 | -43.4 | -48.5 & -41.4 | -43.0 | -48.7 & 30.8 | 23.5 | 18.8 & -32.6 | -38.3 | -48.0 & \hspace{1.25pt} 3.5 | \hspace{1.25pt} 3.8 | \hspace{1.25pt} \textbf{0.6} \\
     &  \checkmark  \checkmark  - -  & -12.7 | -38.5 | -40.4 & -11.9 | -38.7 | -39.8 & -12.3 | -38.6 | -40.1 & \hspace{1.25pt} 2.7 | 18.5 | 10.4 & \hspace{1.25pt} -9.7 | -35.5 | -36.6 & \hspace{1.25pt} 2.5 | \hspace{1.25pt} \textbf{1.5} | \hspace{1.25pt} 2.5 \\
       \hline
       \hline
        \multirow{4}{*}{\rotatebox{90}{HiER} }  &  -  -  - \checkmark  & -34.1 | -42.0 | -47.6 & -35.0 | -42.0 | -47.8 & -34.4 | -42.5 | -47.7 & 24.1 | 22.0 | 17.6 & -27.0 | -35.6 | -46.2 & \hspace{1.25pt} 4.5 | \hspace{1.25pt} 3.5 | \hspace{1.25pt} 0.8 \\
     &  \checkmark  -  - \checkmark  & \hspace{1.25pt} -7.0 | -23.6 | -37.2 & \hspace{1.25pt} \textbf{-6.8} | -22.6 | -36.6 & \hspace{1.25pt} \textbf{-6.9} | -23.1 | -36.9 & \hspace{1.25pt} \textbf{0.0} | \hspace{1.25pt} 4.0 | \hspace{1.25pt} 7.2 & \hspace{1.25pt} -6.1 | -17.8 | -34.5 & \hspace{1.25pt} 0.7 | \hspace{1.25pt} 4.1 | \hspace{1.25pt} 1.9 \\
     &  -  \checkmark  - \checkmark  & -17.2 | -38.2 | -46.8 & -14.5 | -36.6 | -46.7 & -15.7 | -36.8 | -46.7 & \hspace{1.25pt} 7.2 | 18.2 | 16.8 & \hspace{1.25pt} -9.9 | -32.9 | -45.1 & \hspace{1.25pt} 7.6 | \hspace{1.25pt} 5.1 | \hspace{1.25pt} 1.2 \\
     &  \checkmark  \checkmark  - \checkmark  & \hspace{1.25pt} -8.4 | -25.4 | -37.7 & \hspace{1.25pt} -8.3 | -24.9 | -37.3 & \hspace{1.25pt} -8.3 | -25.1 | -37.5 & \hspace{1.25pt} 0.1 | \hspace{1.25pt} 5.4 | \hspace{1.25pt} 7.7 & \hspace{1.25pt} -6.4 | -21.2 | -36.2 & \hspace{1.25pt} 1.2 | \hspace{1.25pt} 3.0 | \hspace{1.25pt} 1.4 \\
       \hline
       \hline
        \multirow{4}{*}{\rotatebox{90}{ISE} }  &  -  -  \checkmark  -  & -14.9 | -37.3 | -41.6 & -15.0 | -37.8 | -41.6 & -15.1 | -37.4 | -41.4 & \hspace{1.25pt} 5.0 | 17.3 | 11.6 & \hspace{1.25pt} -8.3 | -33.6 | -38.8 & \hspace{1.25pt} 3.4 | \hspace{1.25pt} 2.2 | \hspace{1.25pt} 2.0 \\
     &  \checkmark  -  \checkmark  -  & \hspace{1.25pt} -8.1 | -35.9 | -34.1 & \hspace{1.25pt} -8.0 | -35.9 | -34.6 & \hspace{1.25pt} -8.0 | -35.9 | -34.4 & \hspace{1.25pt} \textbf{0.0} | 15.9 | \hspace{1.25pt} 4.3 & \hspace{1.25pt} -6.7 | -32.7 | -27.2 & \hspace{1.25pt} 1.0 | \hspace{1.25pt} 2.0 | \hspace{1.25pt} 3.2 \\
     &  -  \checkmark  \checkmark  -  & -16.3 | -37.3 | -39.1 & -15.7 | -37.5 | -39.0 & -16.1 | -37.5 | -38.9 & \hspace{1.25pt} 6.3 | 17.3 | \hspace{1.25pt} 9.1 & -12.1 | -34.8 | -35.8 & \hspace{1.25pt} 3.2 | \hspace{1.25pt} \textbf{1.5} | \hspace{1.25pt} 2.0 \\
     &  \checkmark  \checkmark  \checkmark  -  & \hspace{1.25pt} -8.1 | -37.6 | -36.1 & \hspace{1.25pt} -7.8 | -37.7 | -36.6 & \hspace{1.25pt} -8.0 | -37.7 | -36.5 & \hspace{1.25pt} \textbf{0.0} | 17.6 | \hspace{1.25pt} 6.1 & \hspace{1.25pt} -6.4 | -34.9 | -31.6 & \hspace{1.25pt} 1.2 | \hspace{1.25pt} \textbf{1.5} | \hspace{1.25pt} 2.1 \\
       \hline
        \multirow{4}{*}{\rotatebox{90}{HiER+} }  &  -  -  \checkmark  \checkmark  & \hspace{1.25pt} -8.8 | -33.4 | -38.1 & \hspace{1.25pt} -8.2 | -33.1 | -38.5 & \hspace{1.25pt} -8.5 | -33.8 | -38.3 & \hspace{1.25pt} 0.3 | 13.4 | \hspace{1.25pt} 8.1 & \hspace{1.25pt} -7.1 | -26.6 | -34.8 & \hspace{1.25pt} 1.7 | \hspace{1.25pt} 3.8 | \hspace{1.25pt} 1.7 \\
     &  \checkmark  -  \checkmark  \checkmark  & \hspace{1.25pt} \textbf{-6.9} | -22.5 | \textbf{-24.2} & \hspace{1.25pt} -7.0 | -22.9 | \textbf{-23.2} & \hspace{1.25pt} -7.0 | -22.8 | \textbf{-23.0} & \hspace{1.25pt} \textbf{0.0} | \hspace{1.25pt} 3.0 | \hspace{1.25pt} \textbf{1.0} & \hspace{1.25pt} \textbf{-5.9} | \textbf{-17.4} | \textbf{-15.0} & \hspace{1.25pt} \textbf{0.6} | \hspace{1.25pt} 3.2 | \hspace{1.25pt} 6.3 \\
     &  -  \checkmark  \checkmark  \checkmark  & \hspace{1.25pt} -8.6 | -34.9 | -34.6 & \hspace{1.25pt} -8.4 | -35.9 | -34.8 & \hspace{1.25pt} -8.5 | -35.3 | -34.6 & \hspace{1.25pt} \textbf{0.0} | 14.9 | \hspace{1.25pt} 4.6 & \hspace{1.25pt} -7.8 | -30.5 | -30.9 & \hspace{1.25pt} 0.7 | \hspace{1.25pt} 2.6 | \hspace{1.25pt} 1.9 \\
     &  \checkmark  \checkmark  \checkmark  \checkmark  & \hspace{1.25pt} -7.7 | \textbf{-21.7} | -33.9 & \hspace{1.25pt} -7.4 | \textbf{-21.5} | -34.4 & \hspace{1.25pt} -7.5 | \textbf{-21.5} | -34.0 & \hspace{1.25pt} \textbf{0.0} | \hspace{1.25pt} \textbf{2.3} | \hspace{1.25pt} 4.1 & \hspace{1.25pt} -7.0 | \textbf{-17.4} | -28.0 & \hspace{1.25pt} \textbf{0.6} | \hspace{1.25pt} 3.1 | \hspace{1.25pt} 3.3 \\
        \end{tabular}
        \vspace{\vspacetable} % Adjust space after table
    \end{table*}
}

\newcommand{\fetchresults}{

\begin{figure*}[p]
    \centering
    \subfloat{%
      \includegraphics[width=\imgsize]{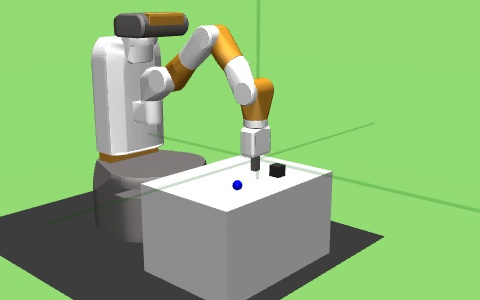}
    }
    \hfill
    \subfloat{%
      \includegraphics[width=\imgsize]{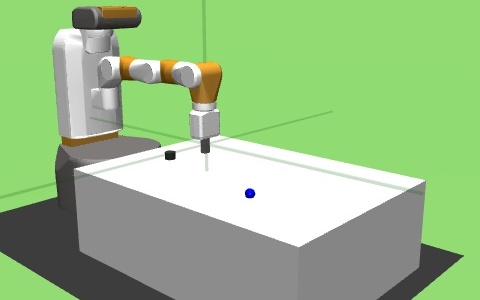}
    }
    \hfill
    \subfloat{%
      \includegraphics[width=\imgsize]{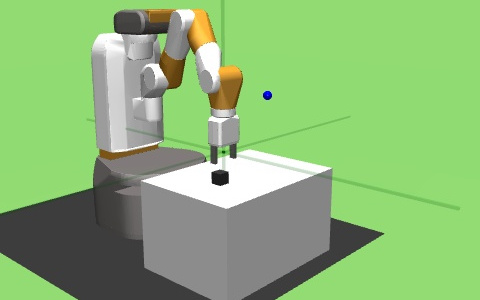}
    }

    \subfloat{%
      \includegraphics[width=\imgsize]{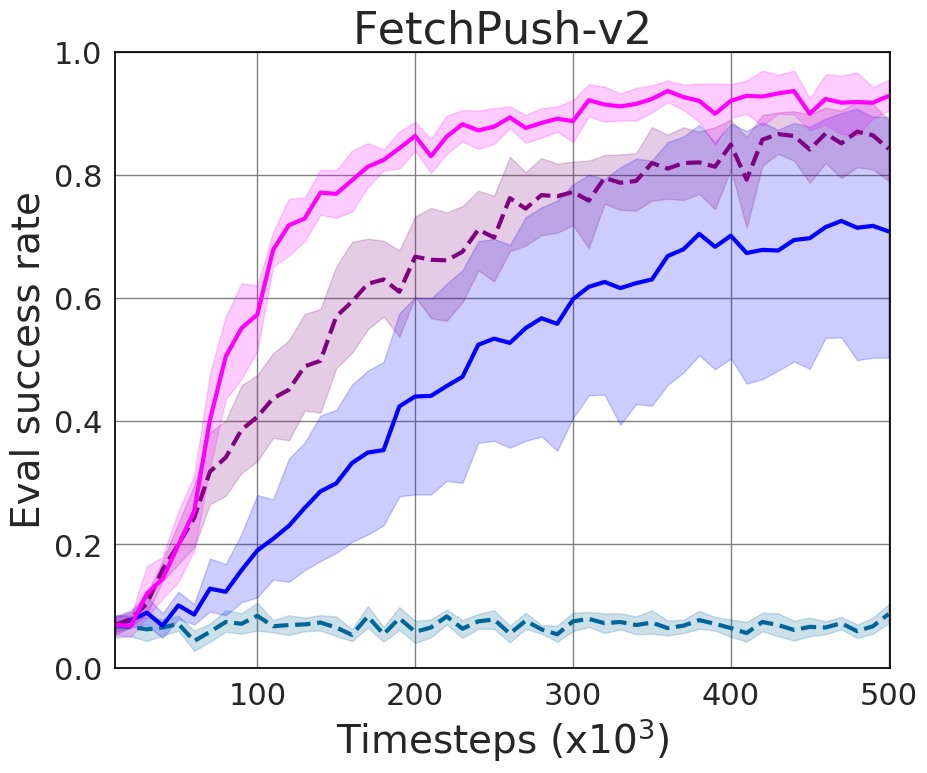}
    }
    \hfill
    \subfloat{%
      \includegraphics[width=\imgsize]{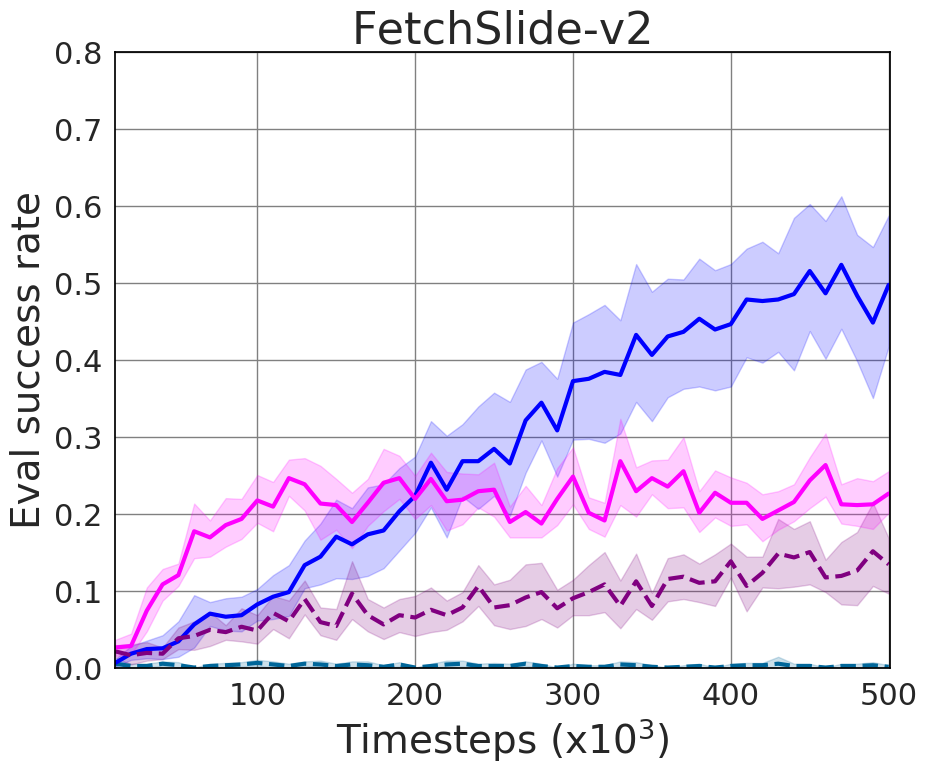}
    }
    \hfill
    \subfloat{%
      \includegraphics[width=\imgsize]{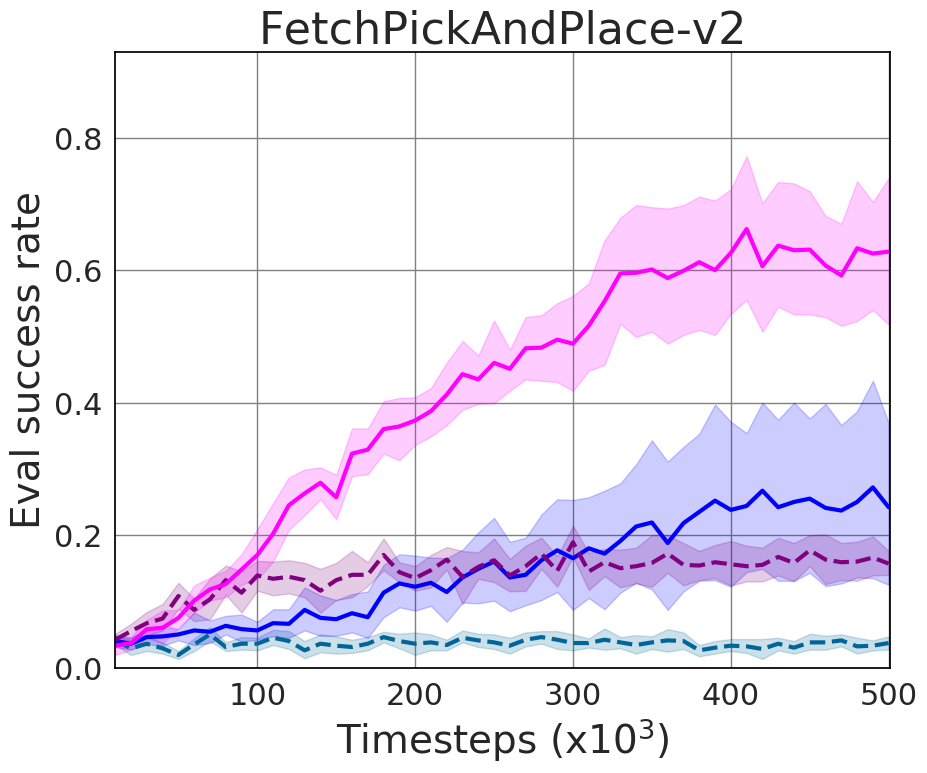}
    }

    \subfloat{%
      \includegraphics[width=\textwidth]{figures/Fetch/FetchLegend_legend.png}
    }

    \caption{Learning curves of HiER compared with its baselines on push, slide, and pick-and-place tasks of the Gymnasium-Robotics Fetch benchmark with 95\% CIs.}
    \label{fig:result_fetch_tasks}
\end{figure*}

\begin{figure*}[p]
\centering
\includegraphics[width=1.0\textwidth]{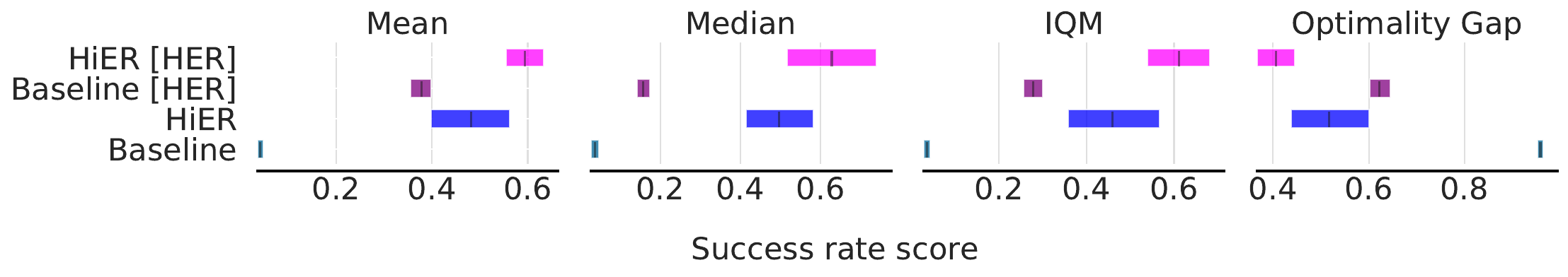}
\caption{Aggregate metrics on the push, slide, and pick-and-place tasks of the Gymnasium-Robotics Fetch benchmark with 95\% CIs. Both HiER (blue) and HiER~[HER] (magenta) significantly outperform the baselines (light blue and purple).}
\label{fig:fetch_agg_metrics}
\end{figure*}

\begin{table*}[p]
    \centering
    \caption{HiER compared to the state-of-the-art based on success rates on push, slide, and pick-and-place tasks of the Gymnasium-Robotics Fetch benchmark. The column-wise best results are marked in bold.}
    \label{tab:results_fetch}
    \begin{tabular}{c|c|cccccc}
          & & \multicolumn{6}{c}{\texttt{FetchPush-v2} | \texttt{FetchSlide-v2} | \texttt{FetchPickAndPlace-v2} } \\
          & \rotatebox{90}{HER} \rotatebox{90}{HiER} & Mean $\uparrow$ & Median $\uparrow$ & IQM $\uparrow$ & OG $\downarrow$ & Max $\uparrow$ & Std $\downarrow$ \\
   \hline
    \multirow{2}{*}{Baselines}  &  - -  & 0.12 | 0.02 | 0.08 & 0.12 | 0.02 | 0.08 & 0.12 | 0.02 | 0.08 & 0.88 | 0.98 | 0.92 & 0.14 | 0.04 | 0.10 & \textbf{0.01} | \textbf{0.01} | \textbf{0.01} \\
 &  \checkmark - & 0.92 | 0.23 | 0.24 & 0.93 | 0.22 | 0.23 & 0.92 | 0.22 | 0.24 & 0.08 | 0.77 | 0.76 & 0.98 | 0.39 | 0.30 & 0.05 | 0.07 | 0.04 \\
 \hline
 \multirow{2}{*}{HiER} &  - \checkmark  & 0.76 | \textbf{0.56} | 0.32 & 0.93 | \textbf{0.55} | 0.17 & 0.83 | \textbf{0.56} | 0.26 & 0.24 | \textbf{0.44} | 0.68 & \textbf{1.00} | \textbf{0.80} | 0.76 & 0.29 | 0.13 | 0.22 \\
 &  \checkmark \checkmark  & \textbf{0.98} | 0.35 | \textbf{0.73} & \textbf{0.99} | 0.36 | \textbf{0.77} & \textbf{0.98} | 0.36 | \textbf{0.73} & \textbf{0.02} | 0.65 | \textbf{0.27} & \textbf{1.00} | 0.39 | \textbf{0.93} & 0.02 | 0.03 | 0.14 \\
    \end{tabular}
\end{table*}

}

\newcommand{\mazeresults}{
    \begin{figure*}[p]
        \centering
        \subfloat[\texttt{PointMaze-Wall-v3}]{%
          \includegraphics[width=.23\linewidth]{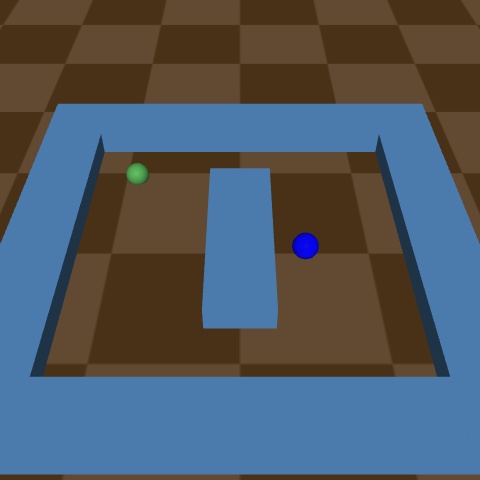}
        }
        \hfill
        \subfloat[Layout of \texttt{PointMaze-Wall-v3}]{%
          \includegraphics[width=.23\linewidth]{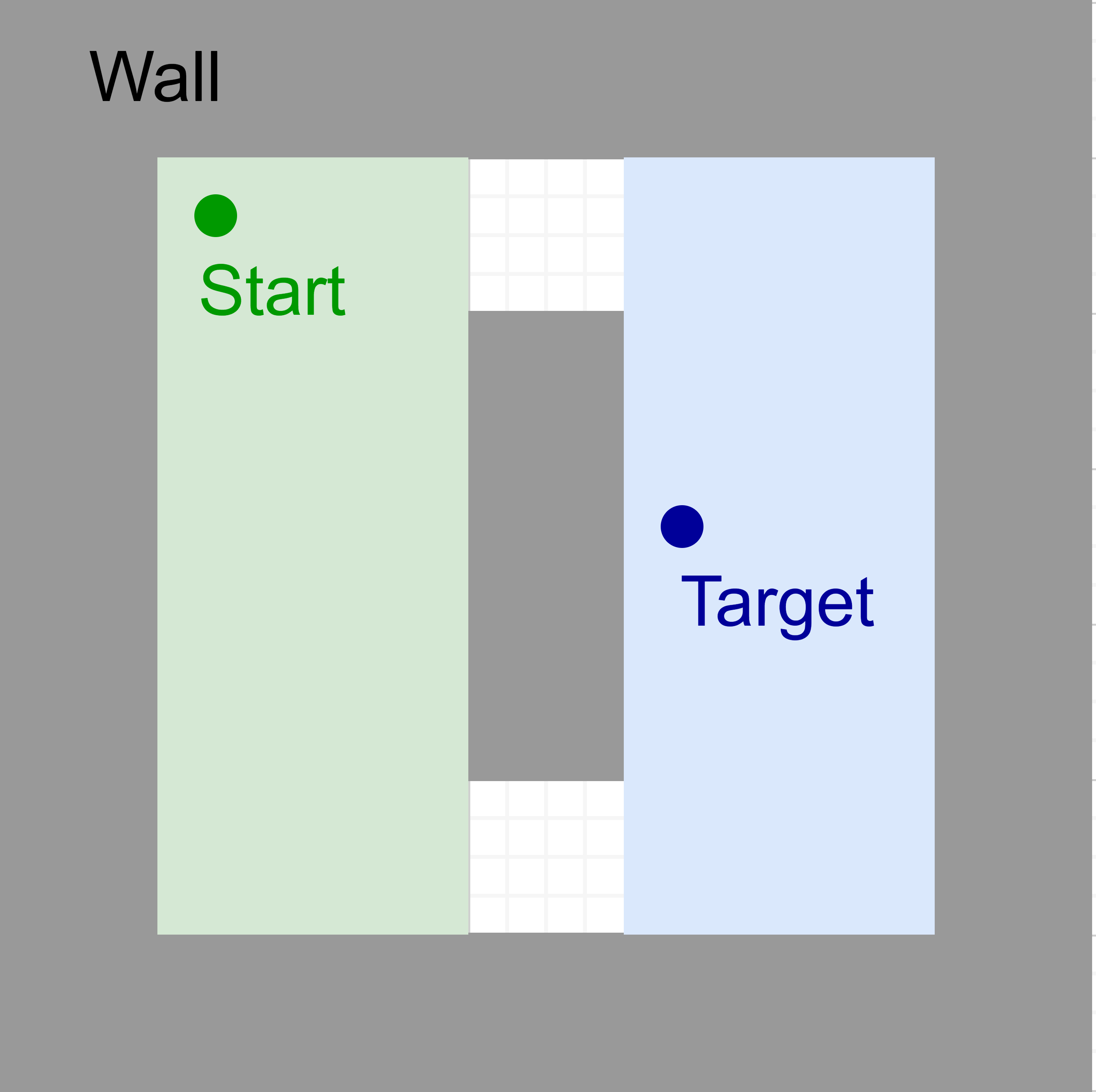}
        }
        \hfill
        \subfloat[\texttt{PointMaze-S-v3}]{%
          \includegraphics[width=.23\linewidth]{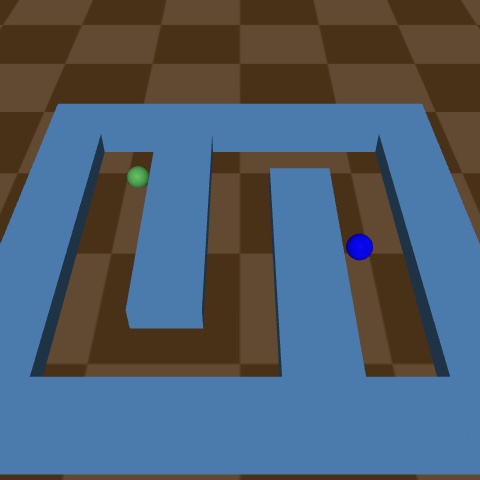}
        }
        \hfill
         \subfloat[Layout of \texttt{PointMaze-S-v3}]{%
          \includegraphics[width=.23\linewidth]{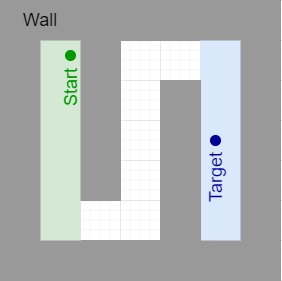}
        }
        \caption{The tasks of Gymnasium-Robotics PointMaze environment~\cite{Maze}. The mazes were custom-made, thus we named them accordingly. The layouts (b) and (d) show the placement of the walls and the possible start and target positions from a top view. The environment is based on the MuJoCo simulator~\cite{mujoco}.}
        \label{fig:result_maze_tasks}
    \end{figure*}

    \begin{figure*}[p]
        \centering
        \subfloat{%
          \includegraphics[width=50mm]{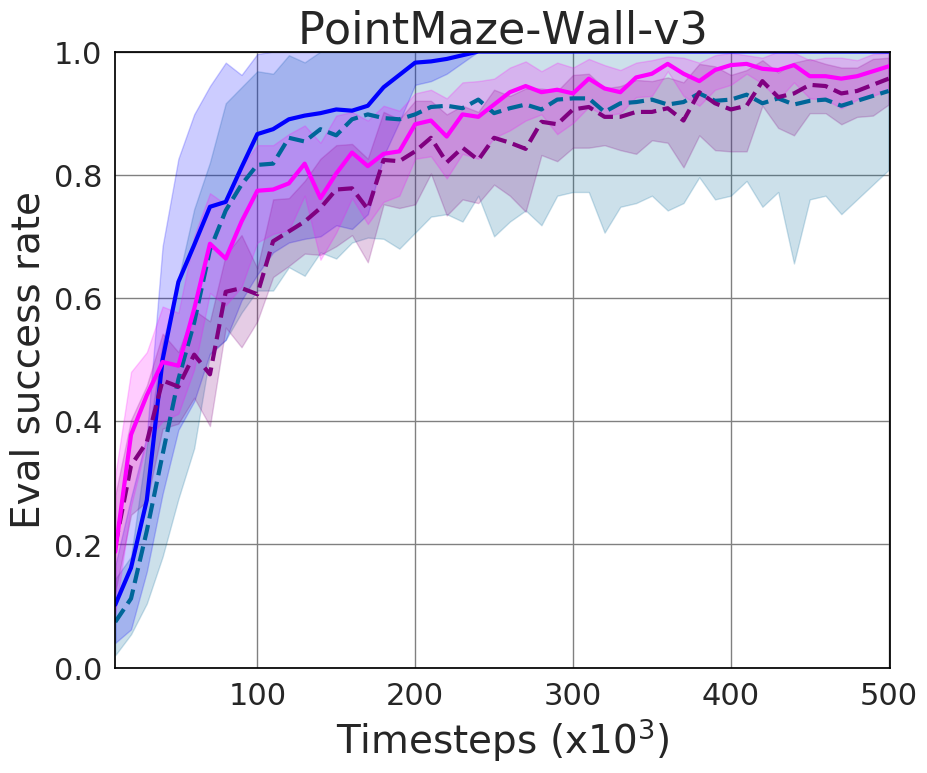}
        }
        \subfloat{%
          \includegraphics[width=50mm]{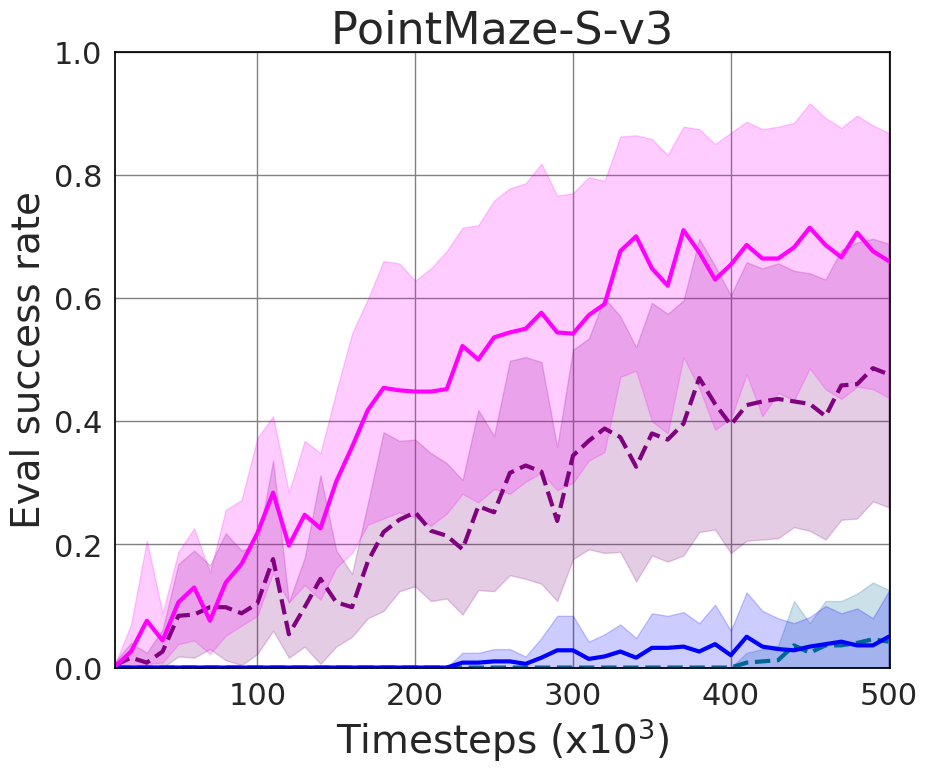}
        }
    
        \subfloat{%
          \includegraphics[width=\textwidth]{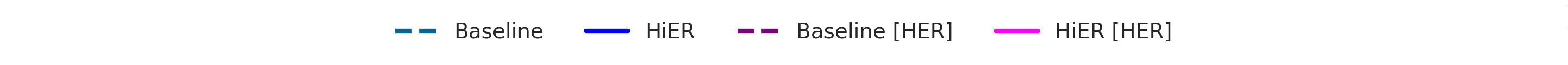}
        }
        
        \caption{Learning curves of HiER compared with its baselines on the Gymnasium-Robotics PointMaze environment with 95\% CIs.}
        \label{fig:result_maze_graphs}
    \end{figure*}
    
    \begin{table*}[p]
        \centering
        \caption{HiER compared to the state-of-the-art based on success rates on the Gymnasium-Robotics PointMaze environment. The column-wise best results are marked in bold.}
       \label{tab:results_maze}
        \begin{tabular}{c|c|cccccc}
              & & \multicolumn{6}{c}{\texttt{PointMaze-Wall-v3} | \texttt{PointMaze-S-v3} } \\
              & \rotatebox{90}{HER} \rotatebox{90}{HiER} & Mean $\uparrow$ & Median $\uparrow$ & IQM $\uparrow$ & OG $\downarrow$ & Max $\uparrow$ & Std $\downarrow$ \\
       \hline
        \multirow{2}{*}{Baselines}  &  - -  & 0.94 | 0.05 & \textbf{1.00} | 0.00 & \textbf{1.00} | 0.00 & 0.06 | 0.95 & \textbf{1.00} | 0.46 & 0.19 | 0.14 \\
     &  \checkmark -  & 0.97 | 0.61 & \textbf{1.00} | 0.76 & 0.99 | 0.69 & 0.03 | 0.39 & \textbf{1.00} | \textbf{1.00} & 0.05 | 0.36 \\
     \hline
     \multirow{2}{*}{HiER} & - \checkmark  & \textbf{1.00} | 0.05 & \textbf{1.00} | 0.00 & \textbf{1.00} | 0.00 & \textbf{0.00} | 0.95 & \textbf{1.00} | 0.28 & \textbf{0.00} | \textbf{0.11} \\
     &  \checkmark \checkmark  & \textbf{1.00} | \textbf{0.80} & \textbf{1.00} | \textbf{0.91} & \textbf{1.00} | \textbf{0.89} & \textbf{0.00} | \textbf{0.20} & \textbf{1.00} | \textbf{1.00} & 0.01 | 0.29 \\
        \end{tabular}
    \end{table*}

}

\newcommand{\otherresults}{

    \begin{figure*}[p]
        \centering
        
        \subfloat[The effect of HiER $\lambda$ versions on the success rate. $\xi$ is fixed at 0.5. ]{%
            \begin{tabular}[b]{@{}c@{}}
                \includegraphics[width=\imgsize]{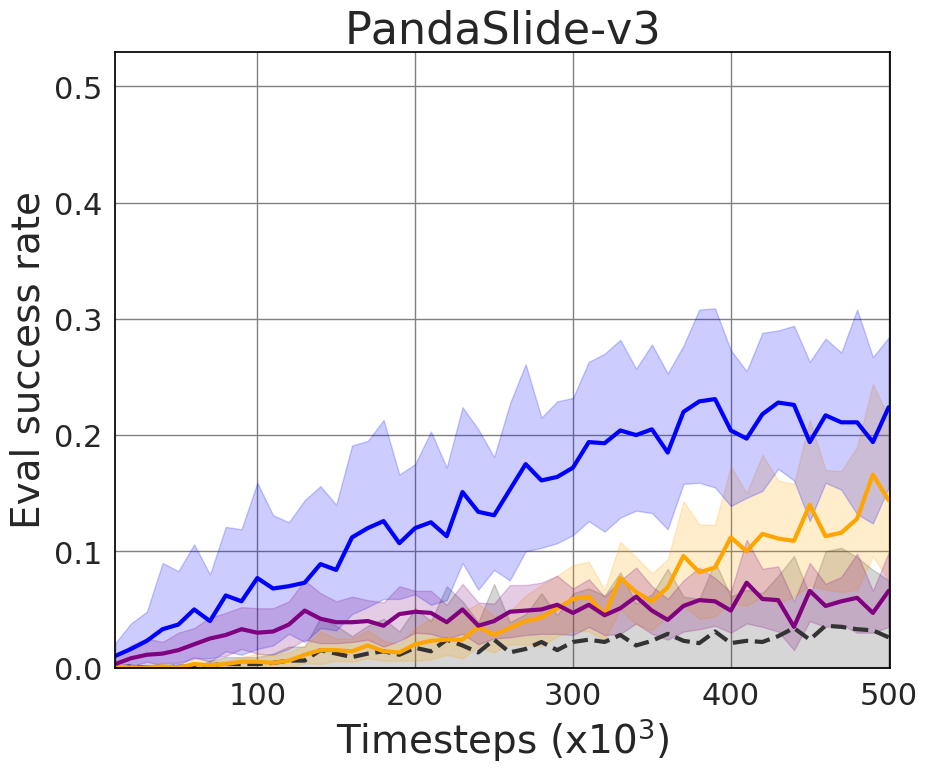} \\
                \includegraphics[width=\imgsize]{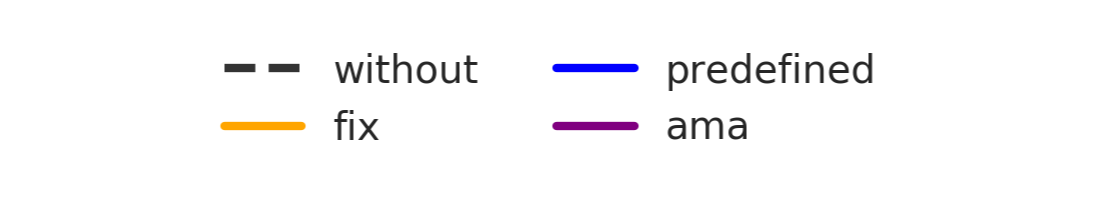}
            \end{tabular}%
        }
        \hfill
        \subfloat[The change of $\lambda$ values over time. $\xi$ is fixed at 0.5.]{%
          \begin{tabular}[b]{@{}c@{}}
                \includegraphics[width=\imgsize]{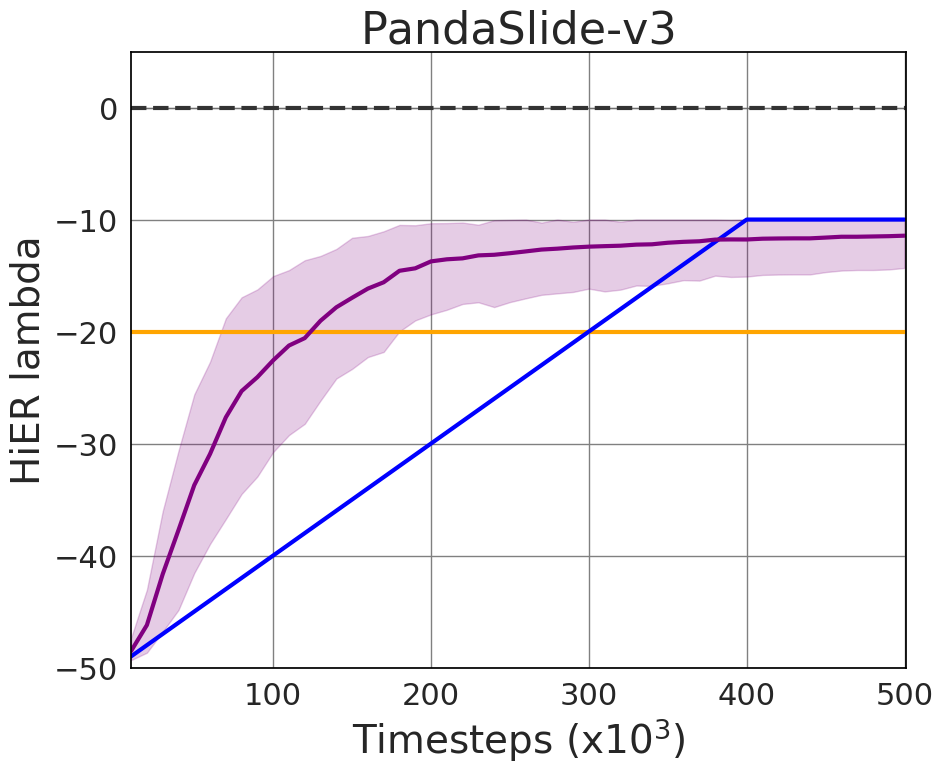} \\
                \includegraphics[width=\imgsize]{figures/HiER/lambda/HiER_lambda_legend.png}
            \end{tabular}%
        }
        \hfill
        \subfloat[The effect of HiER $\xi$ methods. HiER $\lambda$ is set to \texttt{predefined}. ]{%
           \begin{tabular}[b]{@{}c@{}}
                \includegraphics[width=\imgsize]{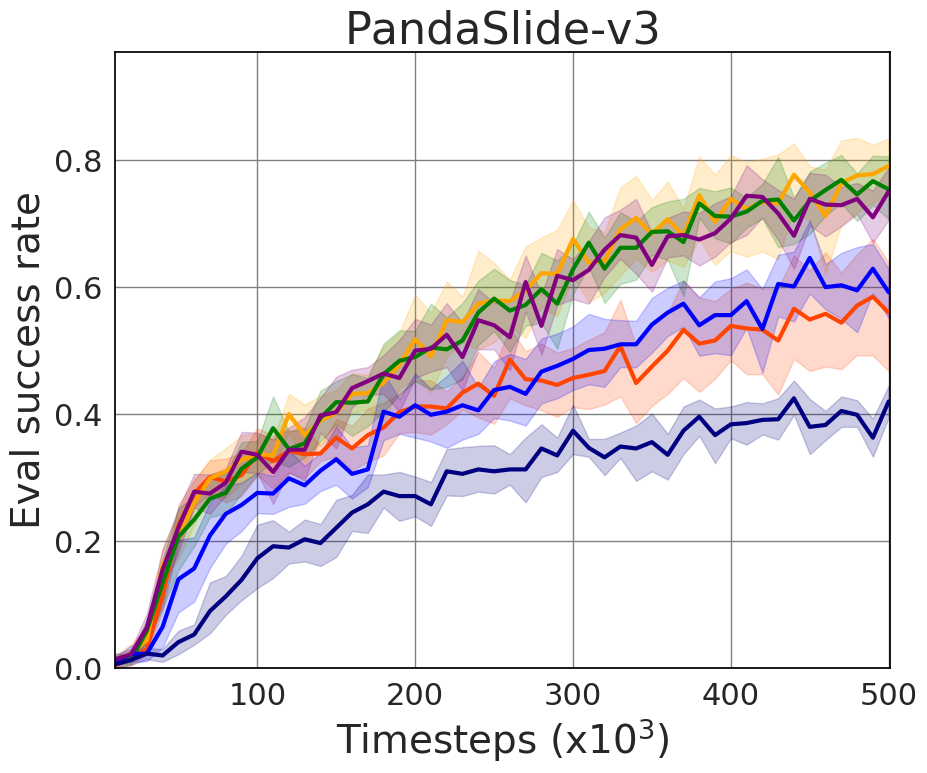} \\
                \includegraphics[width=\imgsize]{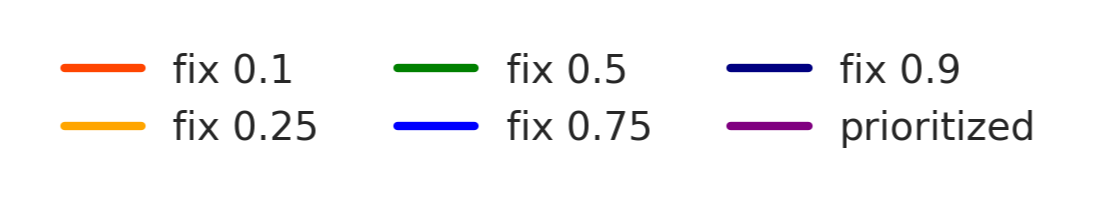}
            \end{tabular}%
        }
        \caption{The analysis of HiER $\lambda$ versions (a) and (b), and HiER $\xi$ versions (c). HiER $\lambda$ \texttt{ama} parameters: $\lambda_0=-50$, $\lambda_{max}=-10$ $M=0$ and $w=20$. The \textit{without} version indicates that HiER was not used.}
        \label{fig:results_hier_modes}
    \end{figure*}
    
    \begin{figure*}[p]
    \centering
    \includegraphics[width=0.8\textwidth]{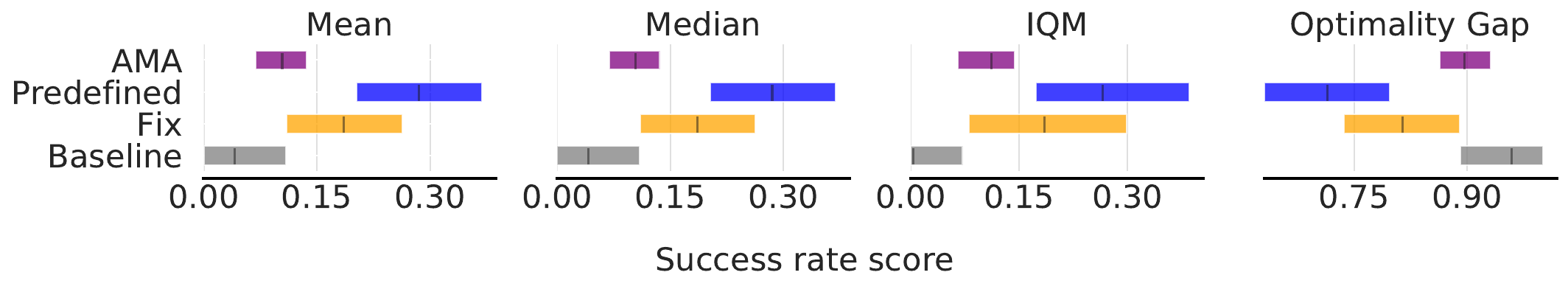}
    \caption{Comparison of different HiER $\lambda$ methods on the slide task of the Panda-Gym benchmark with 95\% CIs. The \texttt{predefined} $\lambda$ method is seemingly superior, although the CIs with the \texttt{fix} $\lambda$ method overlap. HiER $\lambda$ \texttt{ama} parameters: $\lambda_0=-50$, $\lambda_{max}=-10$ $M=0$ and $w=20$. The profiles of HiER $\lambda$ are depicted on Fig.~\ref{fig:results_hier_modes} (b).}
    \label{fig:hier_lambda_agg_metrics}
    \end{figure*}
    
    \begin{figure*}[p]
    \centering
    \includegraphics[width=0.8\textwidth]{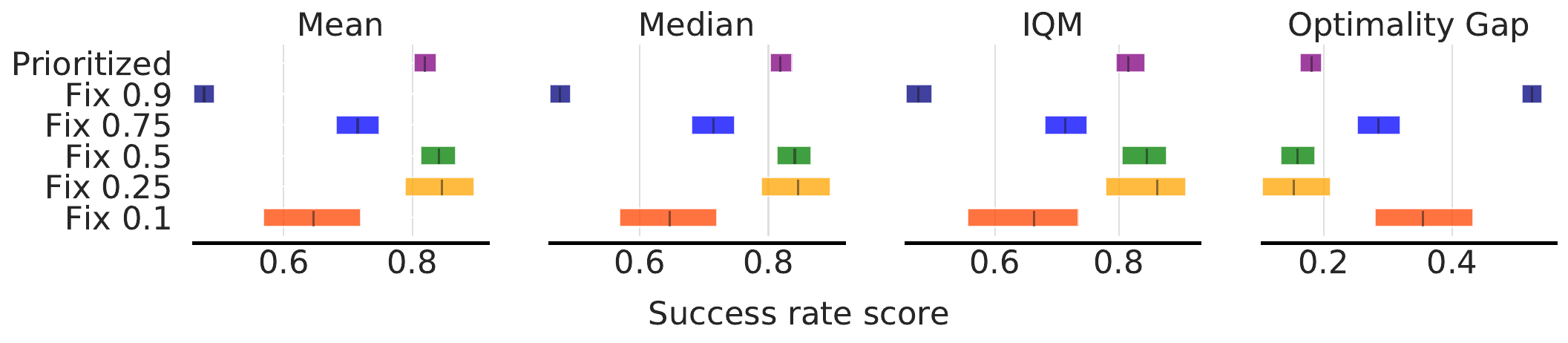}
    \caption{Comparison of different HiER $\xi$ methods on the slide task of the Panda-Gym benchmark with 95\% CIs. The \texttt{fix} $\xi = 0.25$, $\xi = 0.5$, and the \texttt{prioritized} appear to be the best versions in this order, although their CIs overlap.}
    \label{fig:hier_xi_agg_metrics}
    \end{figure*}

    \begin{figure*}[p]
    \centering
    \includegraphics[width=0.9\textwidth]{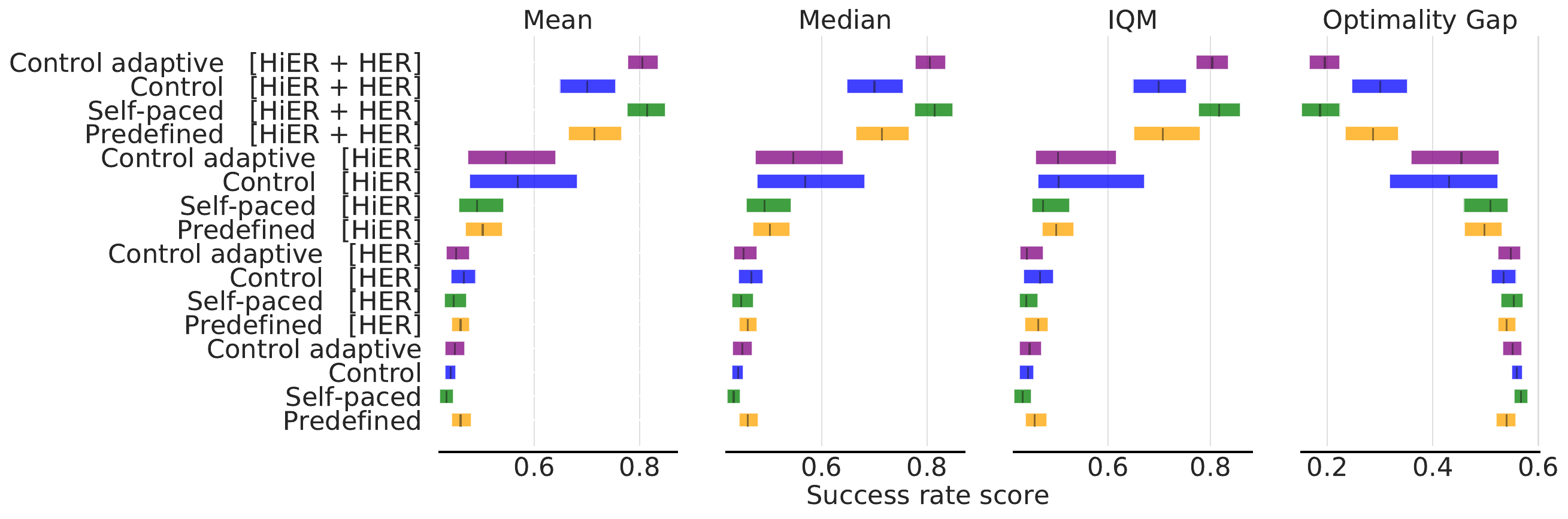}
    \caption{Comparison of different E2H-ISE $c$ methods on the slide task of the Panda-Gym benchmark with 95\% CIs. The parameters of the methods and the point estimates are presented in Tab.~\ref{tab:e2h}.}
    \label{fig:e2h_agg_metrics}
    \end{figure*}

    \begin{table*}[p]
      \centering
          \caption{The effect of the E2H-ISE $c$ methods on the success rates on the \texttt{Panda\-Slide\--v3} task. HiER parameters: $\lambda$ mode \texttt{predefined} and $\xi$ \texttt{fix} with $\xi = 0.5$. E2H-ISE parameters: \texttt{self-paced} $\Psi_{low} = 0.2$, $\Psi_{high} = 0.8$ and $\delta = 0.05$; \texttt{control}: $\psi = 0.8$ and $\delta = 0.01$; \texttt{control} \texttt{adaptive}: $\Delta = 0.2$, $\psi_{max} = 0.9$, and $\delta = 0.01$. The row-wise best results are marked in bold.}
      \label{tab:e2h}
     \begin{tabular}{cc|ccc|ccc|ccc|ccc}
        \multicolumn{2}{c}{Components} & \multicolumn{3}{c}{\texttt{predefined}} & \multicolumn{3}{c}{\texttt{self-paced}} & \multicolumn{3}{c}{\texttt{control}}  & \multicolumn{3}{c}{\texttt{control adaptive}} \\
        
         HER & HiER & Max $\uparrow$ &  Mean $\uparrow$ &  Std $\downarrow$ & Max $\uparrow$ &  Mean $\uparrow$ &  Std $\downarrow$  & Max $\uparrow$ &  Mean $\uparrow$ &  Std $\downarrow$ & Max $\uparrow$ &  Mean $\uparrow$ &  Std $\downarrow$  \\
            \hline
        - &  -  &  \textbf{0.52} & \textbf{0.46} & 0.03 & 0.46 & 0.43 & 0.02 & 0.46 & 0.44 &  0.02 & 0.51 & 0.45 & 0.03  \\
        
        \checkmark &   - &   0.49 & 0.46 & 0.03 & \textbf{0.54} & 0.45& 0.03 & 0.53 & \textbf{0.47} & 0.04 & \textbf{0.54} & 0.45 & 0.04 \\
        
         - &   \checkmark &   0.63 & 0.50 & 0.06 & 0.69 & 0.49 & 0.07 & \textbf{0.95} & \textbf{0.57} & 0.17 &  0.90 & 0.55 & 0.14\\
         
         \checkmark &  \checkmark & 0.85 &  0.71 & 0.08 & \textbf{0.90} & \textbf{0.81} & 0.06 & 0.87 & 0.70 & 0.08 & \textbf{0.90} &  0.80 & 0.05\\

      \end{tabular}
    \end{table*}

    \begin{figure*}[p]
        \centering
        \includegraphics[width=0.8\textwidth]{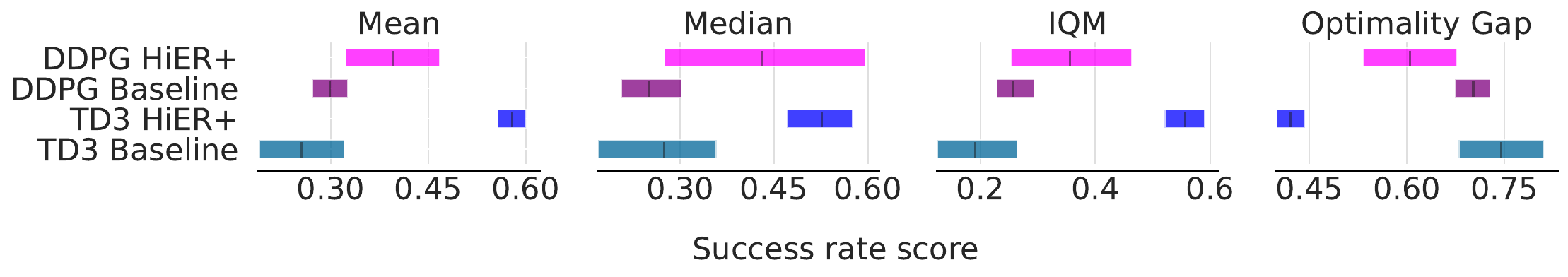}
        \caption{Comparison of the TD3 and DDPG versions of HiER+ with their baselines on the push, slide, and pick-and-place tasks of the Panda-Gym benchmark with 95\% CIs. The point estimates are presented in Tab.~\ref{tab:results_sota_td3_ddpg}.}
        \label{fig:td3_ddpg_agg_metrics}
        \end{figure*}
        
    \begin{table*}[p]
            \centering
            \caption{HiER+ compared to the state-of-the-art based on success rates on the Panda-Gym robotic benchmark in the case of TD3 and DDPG. The column-wise best results for TD3 and DDPG separately are marked in bold.}
          \label{tab:results_sota_td3_ddpg}
            \begin{tabular}{cc|cccccc}
                  & & \multicolumn{6}{c}{\texttt{PandaPush-v3} | \texttt{PandaSlide-v3} | \texttt{PandaPickAndPlace-v3} } \\
                  & RL Algorithm & Mean $\uparrow$ & Median $\uparrow$ & IQM $\uparrow$ & OG $\downarrow$ & Max $\uparrow$ & Std $\downarrow$ \\
             
            \hline
         
              \multirow{2}{*}{DDPG} &  Baseline & 0.25 | 0.56 | 0.08 & 0.23 | 0.56 | 0.08 & 0.24 | 0.56 | 0.08 & 0.75 | 0.44 | 0.92 & 0.42 | 0.74 | 0.11 & \textbf{0.08} | \textbf{0.11} | \textbf{0.01} \\
          
             &  HiER+& \textbf{0.43} | \textbf{0.63} | \textbf{0.13} & \textbf{0.32} | \textbf{0.68} | \textbf{0.12} & \textbf{0.38} | \textbf{0.68} | \textbf{0.12} & \textbf{0.57} | \textbf{0.37} | \textbf{0.87} & \textbf{0.91} | \textbf{0.83} | \textbf{0.20} & 0.27 | 0.22 | 0.03 \\
        
                 \hline
            
            \multirow{2}{*}{TD3}  & Baseline & 0.40 | 0.27 | 0.09 & 0.28 | 0.36 | 0.09 & 0.36 | 0.30 | 0.09 & 0.60 | 0.73 | 0.91 & 0.86 | 0.44 | 0.11 & 0.28 | 0.16 | \textbf{0.01} \\
            
             &  HiER+ & \textbf{0.93} | \textbf{0.53} | \textbf{0.28} & \textbf{0.94} | \textbf{0.56} | \textbf{0.30} & \textbf{0.94} | \textbf{0.54} | \textbf{0.29} & \textbf{0.07} | \textbf{0.47} | \textbf{0.72} & \textbf{0.99} | \textbf{0.63} | \textbf{0.35} & \textbf{0.04} | \textbf{0.08} | 0.05 \\
             
            \end{tabular}
    \end{table*}

}

\begin{document}
\history{Date of publication xxxx 00, 0000, date of current version xxxx 00, 0000.}
\doi{DOI}
\title{HiER: Highlight Experience Replay for Boosting Off-Policy Reinforcement Learning Agents}
\author{\uppercase{Dániel Horváth}\authorrefmark{1,2,3} \IEEEmembership{Member, IEEE},
\uppercase{Jesús Bujalance Martín\authorrefmark{1}}, \uppercase{Ferenc Gábor Erdos\authorrefmark{2}}, \uppercase{Zoltán Istenes\authorrefmark{3}},  and \uppercase{Fabien Moutarde}\authorrefmark{1}.}
\address[1]{Center for Robotics, MINES Paris, PSL University, 75272 Paris, France}
\address[2]{Centre of Excellence in Production Informatics and Control, Institute for Computer Science and Control, Hungarian Research Network, 1111 Budapest, Hungary}
\address[3]{CoLocation Center for Academic and Industrial Cooperation, Eötvös Loránd University, 1117 Budapest, Hungary}
\tfootnote{This work was supported in part by the European Union project RRF-2.3.1-21-2022-00004 within the framework of the Artificial Intelligence National Laboratory and in part by the European Commission through the H2020 project EPIC (https://www.centre-epic.eu/) under grant No. 739592. The work of Dániel Horváth was supported by the Government of France and the Government of Hungary in the framework of "Campus France Bourse du gouvernement français - Bourse Excellence Hongrie".}

\markboth
{Author \headeretal: Preparation of Papers for IEEE TRANSACTIONS and JOURNALS}
{Author \headeretal: Preparation of Papers for IEEE TRANSACTIONS and JOURNALS}

\corresp{Corresponding author: Dániel Horváth (e-mail: daniel.horvath@sztaki.hu).}

\begin{abstract}
Even though reinforcement-learning-based algorithms achieved superhuman performance in many domains, the field of robotics poses significant challenges as the state and action spaces are continuous, and the reward function is predominantly sparse. Furthermore, on many occasions, the agent is devoid of access to any form of demonstration. Inspired by human learning, in this work, we propose a method named highlight experience replay (HiER) that creates a secondary highlight replay buffer for the most relevant experiences. For the weights update, the transitions are sampled from both the standard and the highlight experience replay buffer. It can be applied with or without the techniques of hindsight experience replay (HER) and prioritized experience replay (PER). Our method significantly improves the performance of the state-of-the-art, validated on 8 tasks of three robotic benchmarks. Furthermore, to exploit the full potential of HiER, we propose HiER+ in which HiER is enhanced with an arbitrary data collection curriculum learning method. Our implementation, the qualitative results, and a video presentation are available on the project site: \href{http://www.danielhorvath.eu/hier/}{http://www.danielhorvath.eu/hier/}.
\end{abstract}


\titlepgskip=-15pt

\maketitle

\section{INTRODUCTION}

A high degree of transferability is essential to create universal robotic solutions. While transferring knowledge~\cite{pan_survey_2010,weiss_survey_2016} between domains~\cite{salvato_crossing_2021,barisic_sim2air_2022,horvath_sim2real_obj_2023,horvath_sim2real_mogpe_2023}, robotic systems~\cite{zhou_knowledge_2019}, or tasks~\cite{bao_information-theoretic_2019} is fundamental, it is essential to create and apply universal methods such as reinforcement-learning-based algorithms (RL)~\cite{sutton_reinforcement_2018,naeem_gentle_2020,mohammed_review_2020} which are inspired by the profoundly universal trial-and-error-based human/animal learning.

%transferring skills~\cite{liu_skill_2020}, or learning higher-level concepts~\cite{lazaro-gredilla_beyond_2018}

RL methods, especially combined with neural networks (deep reinforcement learning), were proven to be superior in many fields such as achieving superhuman performance in chess~\cite{silver_mastering_2017-1}, Go~\cite{silver_mastering_2017}, or Atari games~\cite{mnih_human-level_2015}. Nevertheless, in the field of robotics, there are significant challenges yet to overcome. Most importantly, the state and action spaces are continuous which intensifies the challenge of exploration. Oftentimes, discretization is not feasible due to loss of information or accuracy, preventing the application of tabular RL methods with high stability. Furthermore, the reward functions of robotic tasks are predominantly sparse which escalates the difficulty of exploration.

Introducing prior knowledge in the form of reward shaping could facilitate the exploration by guiding the agent toward the desired solution. However, 1) constructing a sophisticated reward function requires expert knowledge, 2) the reward function is task-specific, and 3) the agent might learn undesired behaviors. Another source of prior knowledge could be in the form of expert demonstrations. However, collecting demonstrations is oftentimes expensive (time and resources) or even not feasible. Furthermore, it constrains transferability as demonstrations are task-specific.

In parallel to constructing more efficient RL algorithms such as state-of-the-art actor-critic models (DDPG~\cite{silver_deterministic_nodate,lillicrap_continuous_2019}, TD3~\cite{fujimoto_addressing_2018}, and SAC~\cite{haarnoja_soft_2018}), another line of research focuses on improving existing RL algorithms by controlling the data collection~\cite{mehta_active_2019,luo_accelerating_2020,florensa_reverse_2018,ivanovic_barc_2019,salimans_learning_2018,sukhbaatar_intrinsic_2018,florensa_automatic_2018,pong_skew-fit_2020,racaniere_automated_2020} or the data exploitation~\cite{schaul_prioritized_2016,oh_self-imitation_2018, ferret_self-imitation_2020, wang_boosting_2019, andrychowicz_hindsight_2017,bujalance2023reward} process. Following~\cite{portelas_automatic_2020}, in this work, we consider both the data collection and the data exploitation methods as curriculum learning (CL) methods~\cite{bengio_curriculum_2009,portelas_automatic_2020,wang_survey_2022}. The former is oftentimes referred to as 'traditional' and the latter as 'implicit' CL.

\begin{figure*}[h!tb]
\centering
\includegraphics[width=0.9\textwidth]{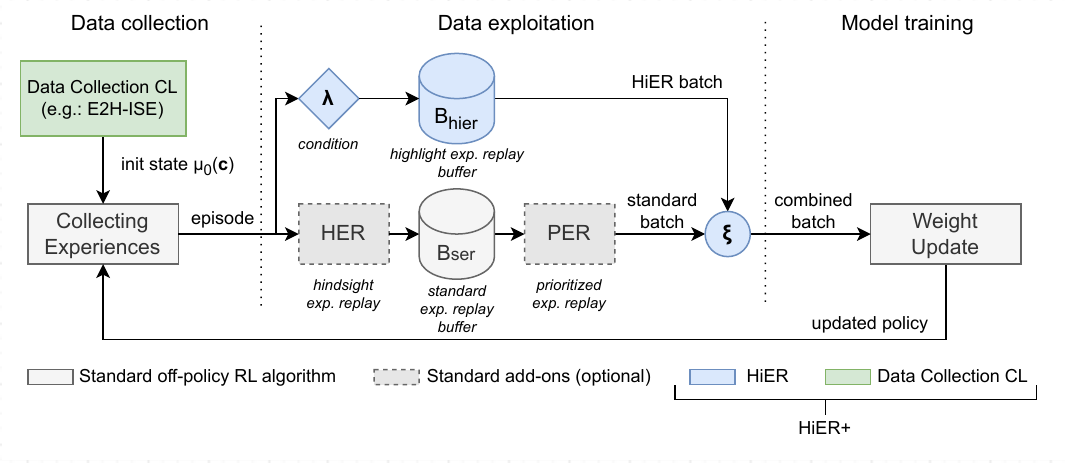}
\hfil
\caption{The overview of HiER and HiER+. For every episode, the initial state is sampled from $\mu_0$. After every episode, the transitions are stored in $\mathcal{B}_{ser}$, and in case the $\lambda$ condition is fulfilled then in $\mathcal{B}_{hier}$ as well. For training, the transitions are sampled from both $\mathcal{B}_{ser}$ and $\mathcal{B}_{hier}$ according to the ratio $\xi$. For a detailed description, see Alg.~\ref{alg:HiER+}.}
\label{fig:HiER+}
\end{figure*}

Our aim is to improve the training of off-policy reinforcement learning agents, particularly in scenarios with continuous state and action spaces, sparse rewards, and the absence of demonstrations. These conditions pose significant challenges for state-of-the-art RL algorithms, due to the challenging problem of exploration. Our main contributions are the following:

\begin{enumerate}
    \item \textbf{HiER}: The highlight experience replay creates a secondary experience replay buffer to store the most relevant transitions. At training, the transitions are sampled from both the standard experience replay buffer and the highlight experience replay buffer. It can be added to any off-policy RL agent and applied with or without the techniques of hindsight experience replay (HER)~\cite{andrychowicz_hindsight_2017} and prioritized experience replay (PER)~\cite{schaul_prioritized_2016}. If only positive experiences are stored in its buffer, HiER can be viewed as a special, automatic demonstration generator as well. Following~\cite{portelas_automatic_2020}, HiER is classified as a data exploitation or implicit curriculum learning method.
    \item \textbf{HiER+}: The enhancement of HiER with an arbitrary data collection (traditional) curriculum learning method. The overview of HiER+ is depicted in Fig.\ref{fig:HiER+}. Furthermore, as an example of the data collection CL method, we propose E2H-ISE, a universal, easy-to-implement \textit{easy2hard} data collection CL method that requires minimal prior knowledge and controls the entropy of the initial state-goal distribution $\mathcal{H}(\mu_0)$ which indirectly controls the task difficulty\footnote{ISE stands for \textit{initial state entropy}.}.
\end{enumerate}

To demonstrate the universality of our methods, HiER is validated on 8 tasks of three different robotic benchmarks~\cite{gallouedec_panda-gym_2021,FetchEnv,Maze} based on two different simulators~\cite{coumans2016pybullet,mujoco}, while HiER+ is evaluated on the push, slide, and pick-and-place tasks of the Panda-Gym~\cite{gallouedec_panda-gym_2021} robotic benchmark. Our methods significantly improve the performance of the state-of-the-art algorithms for each task.

The paper is structured as follows: in Section~\ref{sec:background}~and~\ref{sec:related_works}, the essentials of RL and CL are described followed by a literature review. In Section~\ref{sec:method}~and~\ref{sec:exps}, HiER, E2H-ISE, and HiER+ are presented with the experimental results. Finally, the summary of our findings is provided in Section~\ref{sec:conclusion}.

\section{Background}\label{sec:background}

\subsection{Reinforcement Learning}\label{sec:background_rl}

In reinforcement learning, an agent attempts to learn the optimal policy for a task through interactions with an environment. It can be formalized with a Markov decision process represented by the state space $\mathcal{S}$, the action space $\mathcal{A}$, the transition probability $p(s_{t+1}|s_t,a_t)$, where $s \in \mathcal{S}$ and $a \in \mathcal{A}$, the reward function $r: \mathcal{S} \times \mathcal{A} \rightarrow \mathbb{R}$, the discount factor $\gamma \in [0,1]$, and the initial state distribution $\mu_0$~\cite{sutton_reinforcement_2018}.

Every episode starts by sampling from the initial state distribution $\mu_0$. In every timestep $t \in \mathbb{N}$, the agent performs an action according to its policy $\pi(a|s)$ and receives a reward, a new state\footnote{For simplicity, the environment is considered to be fully observable.}, and a done flag\footnote{Indicating the end of the episode.} $d \in \{0,1\}$ from the environment. In the case of off-policy algorithms, the $(s_t,a_t,s_{t+1},r_t,d_t)$ tuples called transitions are stored in the so-called experience replay buffer $\mathcal{B}_{er}$ which is a circular buffer and the batches for the weight updates are sampled from it.

Learning the optimal policy is formulated as maximizing the expected discounted sum of future rewards or expected return $\mathbb{E}_{s_0}[R_0^{disc}|s_0]$ and $R_t^{disc} = \sum_{i=t}^{T} \gamma^{i-t}r_i$, where $T \in \mathbb{N}$ is the time horizon. Value-based off-policy algorithms learn the optimal policy by learning the optimal $Q$ (action-value) function: $Q^\pi(s_t,a_t) = \mathbb{E}[R_t^{disc}|s_t,a_t]$. 

In multi-goal tasks, there are multiple reward functions $r^g$ parametrized by the goal $g \in \mathcal{G}$. A goal is described with a set of states $\mathcal{S}^g \subset \mathcal{S}$, and it is achieved when the agent is in one of its goal states $s_t \in \mathcal{S}^g$~\cite{florensa_automatic_2018}. Thus, according to~\cite{schaul_universal_2015} and~\cite{florensa_automatic_2018}, the policy is conditioned also on the goal $\pi(a|s,g)$. In our implementation, we simply insert goal $g$ into state $s$ and consequently, when the initial state is sampled from $\mu_0$, the goal is sampled as well. Henceforth, we refer to $\mu_0$ as the initial state-goal distribution.

In robotics, sparse reward function is often formulated as:
 \begin{equation}
    r(s,\cdot) =
    \begin{cases}
      0, & \text{if}\ s \in \mathcal{S}^g \\
      -1, & \text{otherwise}
    \end{cases} \label{eq:r}
  \end{equation}

Another important aspect of an RL task is whether the agent has access to any form of demonstration. A demonstration is an example of the desired (optimal or suboptimal) behavior provided by an external source which can significantly facilitate the exploration~\cite{ramirez_model-free_2022}. Oftentimes, an expert human provides these examples in which case it can be referred to as human demonstrations. Nevertheless, collecting expert demonstrations is expensive and time-consuming, or even not feasible. On the other hand, automatically generating demonstrations presumes that the task can be solved already, which raises the question of why RL training is needed in the first place. For the aforementioned reasons, in this paper, we assume that the agent is devoid of access to any form of demonstration.

\subsection{Curriculum Learning}\label{sec:background_cl}

In this section, the field of CL is briefly presented. For a thorough overview, we refer the reader to~\cite{wang_survey_2022, 
portelas_automatic_2020}.  

CL, introduced by Bengio et al.~\cite{bengio_curriculum_2009}, attempts to facilitate the machine-learning training process. Similar to how humans require a highly-organized training process (introducing different concepts at different times) to become fully-functional adults, machine-learning-based models might as well benefit from a similar type of curriculum.

Originally, the curriculum followed an \textit{easy2hard} or \textit{starting small} structure~\cite{bengio_curriculum_2009}, however, conflicting results with hard example mining~\cite{shrivastava_training_2016} led to a more general definition of CL which did not include the \textit{easy2hard} constraint. 

In supervised learning, a CL framework typically consists of two main components: the difficulty measurer and the training scheduler. The former assigns a difficulty score to the samples, while the latter arranges which samples can be used and when for the weight updates.

According to~\cite{portelas_automatic_2020}, in reinforcement learning, CL can typically control either the data collection or the data exploitation process. The data collection process can be controlled by changing the initial state distribution, the reward function, the goals, the environment, or the opponent. The data exploitation process can be controlled by transition selection or transition modification. HiER belongs to the data exploitation branch of CL while E2H-ISE is classified as a data collection CL method.

\subsection{Evaluation methods}

State-of-the-art deep reinforcement learning models are compared based on just a few experiments, primarily due to constraints on training time. Therefore, simple point estimates of aggregate performance such as mean and median scores across tasks are insufficient as they do not capture the statistical uncertainty implied by the finite number of training runs. In this section, we present the most relevant statistical evaluation methods utilized in RL.

In general, confidence intervals (CIs) are beneficial to measure uncertainty. The bootstrap CI method creates multiple datasets by resampling with replacement from a set of data points (results of independent training runs). As the distribution of the means of the resampled datasets approaches a normal distribution\footnote{Central limit theorem.}, the CI can be calculated. Traditionally, bootstrap CI is performed on a single task~\cite{chan_measuring_2020,colas_how_2018,henderson_deep_2018}. Agarwal et al.~\cite{agarwal_deep_2021} proposes the method of stratified bootstrap CI which performs a bootstrap CI across multiple tasks using stratified sampling.

Another useful evaluation method is presenting the performance profiles. A tail distribution function is defined as $F(\tau) = P(X > \tau)$, where $\tau \in \mathbb{R}$, and $X$ is a real-valued random variable\footnote{Performance estimates are random variables, based on a finite number of runs.}. The performance profiles are beneficial for comparing different algorithms at a glance. In mathematical terms, $X$ has stochastic dominance over $Y$ if $P(X > \tau) \geq P(Y > \tau)$, for all $\tau$, and for some $\tau$ $P(X > \tau) > P(Y > \tau)$, where $X$ and $Y$ are random variables.  Two main versions are the run-score distribution~\cite{agarwal_deep_2021} and the average-score distributions~\cite{bellemare_arcade_2013}. Examples of performance profiles are presented in Fig.~\ref{fig:all_performance_profile} and the left side of Fig.~\ref{fig:panda_performance_profile_and_prob}. 

%Agarwal et al.~\cite{agarwal_deep_2021} improved the approach of Dolan and Moré~\cite{dolan_benchmarking_2002} with an uncertainty estimate based on stratified bootstrapping.

Displaying the probability of improvement is another beneficial evaluation method. It shows the probability of Algorithm $X$ exceeding Algorithm $Y$ in a set of tasks. Important to note that it only indicates the probability of improvement and not the magnitude of the improvement.

Finally, standard aggregate performance metrics have shortcomings. The median has high variability and it is unchanged even when half of the results are zero, while the mean can be significantly influenced by some outliers. Thus,~\cite{agarwal_deep_2021} proposes the interquartile mean (IQM) and the optimality gap (OG) as alternatives to the median and the mean. IQM removes the bottom and top 25\% of the runs and calculates the mean of the remaining 50\% of the runs. The OG represents the shortfall of the algorithm in achieving a desirable target. It is important to note that the extent to which an algorithm surpasses the desired target does not affect its OG score.

\section{Related Works}\label{sec:related_works}

The summary of the related works is presented in Tab.~\ref{tab:related_works_comaprison}. Following Section~\ref{sec:background_cl}, the CL algorithms are categorized as data exploitation or data collection methods, presented in Section~\ref{sec:related_works_data_exploitation}~and~\ref{sec:related_works_data_collection}. Data exploitation methods either modify the transitions or control the transition selection. The structure and the performance of HiER are compared with the state-of-the-art in Section~\ref{sec:method_hier}~and~\ref{sec:exps}. On the other hand, the data collection methods presented in this section, either control the initial state distribution or the goal distribution. The E2H-ISE method controls both the initial state distribution and the goal distribution as the state space is augmented with the goal, as described in Section~\ref{sec:background_rl}. Comparison with the state-of-the-art is presented in Section~\ref{sec:method_cl}~and~\ref{sec:exps}.

{
\renewcommand{\arraystretch}{1.5} % Adjust row spacing here

\begin{table*}[h!tb]
        \centering
        \caption{Summary of related works.}
        \label{tab:related_works_comaprison}
        \begin{tabular}{c|c|c|p{6cm}}
        Type of CL  & What does CL control? &  Work  & Short description \\
        \hline

         \multirow{12}{*}{\raisebox{6.0ex}{Data exploitation}} & \multirow{4}{*}{\raisebox{2.0ex}{Transition modification}} & \multirow{2}{*}{\raisebox{1.0ex}{HER~\cite{andrychowicz_hindsight_2017}}} & Creating virtual episodes by changing the desired goal to the achieved goal. \\ 
        \cline{3-4}
        &  & \multirow{2}{*}{\raisebox{1.0ex}{R\textsuperscript{2}~\cite{bujalance2023reward}}}  &  Reward relabelling of last transitions of successful episodes. \\
        \cline{2-4}
        
        &  \multirow{8}{*}{\raisebox{4.0ex}{Transition selection}} & \multirow{2}{*}{\raisebox{1.0ex}{PER~\cite{schaul_prioritized_2016}}} & Transition selection is based on the last TD-error of the given transition. \\
        \cline{3-4}
        &  & \multirow{2}{*}{\raisebox{1.0ex}{SIL~\cite{oh_self-imitation_2018}}} & Transition selection is based on the clipped advantage. \\
        \cline{3-4}
        &  & \multirow{2}{*}{\raisebox{1.0ex}{ERE~\cite{wang_boosting_2019}}}  & In transition selection, recent data is prioritized, without forgetting the old transitions. \\ 
        \cline{3-4}
        &  & \multirow{2}{*}{\raisebox{1.0ex}{\textbf{HiER (ours)}}} & \textbf{Creating a secondary replay buffer for the most relevant experiences.} \\
        \hline
        
        \multirow{16}{*}{\raisebox{8.0ex}{Data collection}} & \multirow{6}{*}{\raisebox{3.0ex}{Initial state distribution}} & \multirow{2}{*}{\raisebox{1.0ex}{Reverse curriculum generation~\cite{florensa_reverse_2018}}} & Inititial state is close to the goal. The distance to the goal is gradually increased throughout the training. \\
        \cline{3-4}
        & & \multirow{2}{*}{\raisebox{1.0ex}{BaRC~\cite{ivanovic_barc_2019}}} & Generalization of~\cite{florensa_reverse_2018} by approximating backward reaching sets. \\
        \cline{3-4}
        
        & & \multirow{2}{*}{\raisebox{1.0ex}{Salimans and Chen~\cite{salimans_learning_2018}}} & Initial state is sampled from human demonstration. The distance to the goal is gradually increased. \\
        \cline{2-4}
       
        &  \multirow{8}{*}{\raisebox{4.0ex}{Goals}} &  \multirow{2}{*}{\raisebox{1.0ex}{Asymmetric self-play~\cite{sukhbaatar_intrinsic_2018}}} & Training an agent against differently capable versions of itself to enhance its adaptability and robustness. \\ 
        \cline{3-4}

         & & \multirow{2}{*}{\raisebox{1.0ex}{Goal GAN~\cite{florensa_automatic_2018}.}} & A generator is trained to output new goals with appropriate (intermediate) difficulty. \\
         \cline{3-4}

        & & \multirow{2}{*}{\raisebox{1.0ex}{Skew-Fit~\cite{pong_skew-fit_2020}}} & Maximizing the entropy of the goal-conditioned visited states by giving higher weights to rare samples. \\
        \cline{3-4}

        & & \multirow{2}{*}{\raisebox{1.0ex}{Racani\`{e}re et al.~\cite{racaniere_automated_2020}.}}  & Setter agent generates goals for the solver agent considering goal validity, feasibility, and coverage. \\
        \cline{2-4}

        & \multirow{2}{*}{\raisebox{1.0ex}{Initial state-goal dist.}} & \multirow{2}{*}{\raisebox{1.0ex}{\textbf{E2H-ISE (ours)}}}   &  \textbf{The initial state-goal distribution entropy is gradually increased throughout the training.} \\
        \end{tabular}
    \end{table*}
}

\subsection{Data exploitation}\label{sec:related_works_data_exploitation}

Schaul et al.~\cite{schaul_prioritized_2016} proposed the technique of prioritized experience replay (PER) which controls the transition selection by assigning priority (importance) scores to the samples of the replay buffer based on their last TD error~\cite{sutton_learning_1988} and thus, instead of uniformly, they are sampled according to their priority. Additionally, as high-priority samples would bias the training, importance sampling is applied. 

As a form of prioritization, Oh et al.~\cite{oh_self-imitation_2018} introduced self-imitation learning (SIL) for on-policy RL. The priority is computed based on the clipped advantage. Furthermore, the technique of clipped advantage is utilized to incentivize positive experiences. By modifying the Bellmann optimality operator, Ferret et al.~\cite{ferret_self-imitation_2020} introduced self-imitation advantage learning which is a generalized version of SIL for off-policy RL.

Wang et al.~\cite{wang_boosting_2019} presented the method of emphasizing recent experience (ERE) which is a transition selection technique for off-policy RL agents. It prioritizes recent data without forgetting the past while ensuring that updates of new data are not overwritten by updates of old data.

Andrychowicz et al.~\cite{andrychowicz_hindsight_2017} introduced the technique of hindsight experience replay (HER) which performs transition modification to augment the replay buffer by adding virtual episodes. After collecting an episode and adding it to the replay buffer, HER creates virtual episodes by changing the (desired) goal to the achieved goal at the end state (or to another state depending on the strategy) and relabeling the transitions before adding them to the replay buffer. 

Bujalance and Moutarde~\cite{bujalance2023reward} propose reward relabeling to guide exploration in sparse-reward robotic environments by giving bonus rewards for the last $L$ transitions of the episodes.

\subsection{Data collection}\label{sec:related_works_data_collection}

Florensa et al.~\cite{florensa_reverse_2018} presented the reverse curriculum generation method to facilitate exploration for model-free RL algorithms in sparse-reward robotic scenarios. At first, the environment is initialized close to the goal state. For new episodes, the distance between the initial state and the goal state is gradually increased. As prior knowledge, at least one goal state is required. To sample 'nearby' feasible states, the environment is initialized in a certain seed state (in the beginning at a goal state), and then, for a specific time, random Brownian motion is executed.

Ivanovic et al.~\cite{ivanovic_barc_2019} proposed the backward reachability curriculum method (BaRC) which is a generalization of~\cite{florensa_reverse_2018} utilizing prior knowledge of the simplified, approximate dynamics of the system. They compute the approximate backward reaching sets using the Hamilton-Jacobi reachability formulation and sample from them using rejection sampling. 

Salimans and Chen~\cite{salimans_learning_2018} facilitate exploration by utilizing one human demonstration. In their method, the initial states come from the demonstration. More precisely, until a timestep $t_{D} \in \mathbb{N}$, the agent copies the actions of the demonstration, and after $t_{D}$, it takes actions according to its policy. During the training, $t_{D}$ is moved from the end of the demonstration to the beginning of the demonstration. Their method outperformed state-of-the-art methods in the Atari game Montezuma’s Revenge. Nevertheless, arriving at the same state after a specific sequence of actions (as in the demonstration) is rather unlikely, especially when the transition function is profoundly stochastic, such as in robotics.

Sukhbaatar et al.~\cite{sukhbaatar_intrinsic_2018} present automatic curriculum generation with asymmetric self-play of two versions of the same agent. One proposes tasks for the other to complete. With an appropriate reward structure, they automatically create a curriculum for exploration.

Florensa et al.~\cite{florensa_automatic_2018} create a curriculum for multi-goal tasks by sampling goals of intermediate difficulty (Goal GAN). First, the goals are labeled based on their difficulty, and then a generator is trained to output new goals with appropriate difficulty to efficiently train the agent.

Pong et al.~\cite{pong_skew-fit_2020} proposed Skew-Fit, an automatic curriculum that attempts to create a better coverage of the state space by maximizing the entropy of the goal-conditioned visited states $\mathcal{H}(\mathcal{S}|\mathcal{G})$ by giving higher weights to rare samples. Skew-Fit converges to uniform distribution under specific conditions.

Racani\`{e}re et al.~\cite{racaniere_automated_2020} proposed an automatic curriculum generation method for goal-oriented RL agents by training a setter agent to generate goals for the solver agent considering goal validity, goal feasibility, and goal coverage.

The data collection CL methods are relatively disparate, however, some share specific characteristics. The methods that control the initial state distribution~\cite{florensa_reverse_2018, ivanovic_barc_2019, salimans_learning_2018} attempt to reduce the task difficulty by proposing less challenging starting positions. Other algorithms~\cite{sukhbaatar_intrinsic_2018, racaniere_automated_2020}, utilize a secondary agent to train the protagonist. Instead of focusing on task difficulty, the E2H-ISE algorithm controls the entropy of the init-goal state distribution $\mathcal{H}(\mu_0)$. Among the considered methods, only Skew-Fit~\cite{pong_skew-fit_2020} controls the entropy but in that case, it is the entropy of the goal-conditioned visited states $\mathcal{H}(\mathcal{S}|\mathcal{G})$, not $\mathcal{H}(\mu_0)$.

\section{Method}\label{sec:method}

In this Section, our contributions are presented. First, HiER in Section~\ref{sec:method_hier}, and then E2H-ISE and HiER+ in Section~\ref{sec:method_cl} and ~\ref{sec:method_HiER+}. Our implementation is available at our git repository\footnote{ \href{https://github.com/sztaki-hu/hier}{https://github.com/sztaki-hu/hier}}.

\subsection{HiER}\label{sec:method_hier}

Humans remember certain events stronger than others and tend to replay them more frequently than regular experiences thus learning better from them~\cite{kumaran_what_2016}. As an example, an encounter with a lion or scoring a goal at the last minute will be engraved in our memory. Inspired by this phenomenon, HiER attempts to find these events and manage them differently than regular experiences. In this paper, only positive experiences are considered with HiER, thus it can be viewed as a special, automatic demonstration generator as well.

PER and HER control what transitions to store in the experience replay buffer and how to sample from them. Contrary to them, HiER creates a secondary experience replay buffer. Henceforth, the former buffer is called standard experience replay buffer $\mathcal{B}_{ser}$, and the latter is referred to as highlight experience replay buffer $\mathcal{B}_{hier}$. At the end of every episode, HiER stores the transitions in $\mathcal{B}_{hier}$ if certain criteria are met. For updates, transitions are sampled both from the $\mathcal{B}_{ser}$ and $\mathcal{B}_{hier}$ based on a given sampling ratio. HiER is depicted in Fig.~\ref{fig:HiER+} marked in blue.

The criteria can be based on any type of performance measure, in our case, the undiscounted sum of rewards $R = \sum_{i=0}^{T}r_i$ was chosen. The reward function $r$ is formulated as in Eq.~(\ref{eq:r}). Although more complex criteria are possible, in this work, we consider only one performance measure and one criterion: if $R$ is greater than a threshold $\lambda \in \mathbb{R}$ then all the transitions of that episode are stored in $\mathcal{B}_{hier}$ and $\mathcal{B}_{ser}$, otherwise only in $\mathcal{B}_{ser}$. Nevertheless, $\lambda$ can change in time, thus we define a $\lambda_j$ for every $j$ where $j \in \mathbb{N}$ is the index of the episode. In this work, the following $\lambda$ modes were considered:
\begin{itemize}
    \item \texttt{fix}: $\lambda_j = Z_\lambda$ for every $j$ where $Z_\lambda \in \mathbb{R}$ is a constant.\footnote{We also tried a version with $n$ highlight buffers and $n$ thresholds $Z_1, Z_2, \ldots, Z_n$. An episode is stored in the highlight buffer with the highest $Z_i$ for which $ R > Z_i $.}
    
    \item \texttt{predefined}: $\lambda$ is updated according to a predefined profile. Profiles could be arbitrary, such as linear, square-root, or quadratic. In this work, only the linear profile with saturation was considered:
    \begin{equation}
         \lambda_{j} = \min \left( 1,\ \frac{t}{T_{total} \cdot z_{sat}} \right) \label{eq:lambda_linear} 
    \end{equation}
    where $t \in \mathbb{N}$ and $T_{total} \in \mathbb{N}$ are the actual, and the total timesteps of the training and $z_{sat} \in [0,1]$ is a scaler, indicating the start of the saturation.\footnote{In the equation, $\lambda_j$ does not directly depend on $j$. However as $t$ increases, so does $j$ and $\lambda_j$ with it.}
    
    \item \texttt{ama} (adaptive moving average): $\lambda$ is updated according to:
    \begin{equation}
    \lambda_j =
    \begin{cases}
       \min \left( \lambda_{max},\ M + \frac{1}{w} \sum_{i=1}^{w}R_{j-i}\right), & \text{if}\ j > w \\
       \lambda_0, & \text{otherwise}
      \label{eq:lambda_ama}
    \end{cases}
    \end{equation}
    
    where $\lambda_0 \in \mathbb{R}$ is the initial value of $\lambda$, while $\lambda_{max} \in \mathbb{R}$ is the maximum value allowed for $\lambda$. Furthermore, $w \in \mathbb{Z}^+$ is the window size and $M \in \mathbb{R}$ is a constant shift.\footnote{In an alternative version $M$ is not a constant but relative to $\frac{1}{w} \sum_{i=1}^{w}R_{j-i}$.} 
\end{itemize}

Another relevant aspect of HiER is the sampling ratio between $\mathcal{B}_{ser}$ and $\mathcal{B}_{hier}$ for weight update, defined by $\xi \in [0,1]$. It can change in time, updated after every weight update, thus we define a $\xi_k$ for every $k$ where $k \in \mathbb{N}$ is the index of the weight update. The following versions were considered:
\begin{itemize}
    \item \texttt{fix}: $\xi_k = Z_\xi$ for every $k$ where $Z_\xi \in \mathbb{R}$ is a constant.
    \item \texttt{prioritized}: $\xi$ is updated according to:\footnote{Similarly as in the case of PER.}
    \begin{gather}
        \xi_k =  \frac{L_{hier,k}^{\alpha_p}}{L_{hier,k}^{\alpha_p} + L_{ser,k}^{\alpha_p}} \label{eq:xi}  
    \end{gather}
    where $L_{hier,k} \in \mathbb{R}$ and $L_{ser,k} \in \mathbb{R}$ are the TD errors of the training batches sampled from $\mathcal{B}_{hier}$ and $\mathcal{B}_{ser}$ at $k$. The parameter $\alpha_p \in [0,1]$ determines how much prioritization is used.\footnote{If $\alpha_p = 0$, then $\xi = 0.5$ regardless $L_{hier,k}$ and $L_{ser,k}$.}
\end{itemize}

Sampling from $\mathcal{B}_{hier}$ and not only from $\mathcal{B}_{ser}$ introduce a bias towards the experiments collected in $\mathcal{B}_{hier}$. This bias is similar in nature to the case when demonstrations are utilized. In that scenario, the expert demonstrations are sampled and combined with online experience, biasing the exploration towards the desired behavior. In our case, as the agent is devoid of any form of demonstration, $\mathcal{B}_{hier}$ serves similarly as a demonstration buffer. This bias is essential for achieving enhanced performance (presented in Section~\ref{sec:exps}). However, some characteristics of the proposed methods mitigate the sampling bias. The \texttt{predefined} and the \texttt{ama} $\lambda$ methods alleviate the bias by setting the entry of $\mathcal{B}_{hier}$ lower at the beginning and gradually increasing it resulting in a higher cardinality for $\mathcal{B}_{hier}$ and higher similarity between $\mathcal{B}_{hier}$ and $\mathcal{B}_{ser}$. Furthermore, the presented \texttt{prioritized} $\xi$ method prevents overfitting on the data of $\mathcal{B}_{hier}$ as low $L_{hier}$ loss reduces $\xi$ (see Eq.~(\ref{eq:xi})). On the other hand, the bias could be further reduced by gradually decreasing $\xi$ over time, or the gradient of the data from $\mathcal{B}_{hier}$ could be scaled, similarly to importance sampling in the case of PER~\cite{schaul_prioritized_2016}. 

%Nevertheless, our experimental results presented in Section~\ref{sec:exps} indicate that the remaining bias does not affect considerably the results as our methods significantly outperform the state-of-the-art.

Another relevant aspect worth detailing is the difference between the \texttt{prioritized} $\xi$ method and PER. While PER changes the probability distribution of selecting specific transitions from $\mathcal{B}_{ser}$ based on their individual TD error, the \texttt{prioritized} $\xi$ method controls sampling between $\mathcal{B}_{ser}$ and $\mathcal{B}_{hier}$ based on the mean TD error of the data selected from $\mathcal{B}_{ser}$ and $\mathcal{B}_{hier}$. Thus, the sampling distribution of PER has $|\mathcal{B}_{ser}|$ outputs while the sampling distribution of the \texttt{prioritized} $\xi$ method has two outputs, one for $\mathcal{B}_{ser}$ and one for $\mathcal{B}_{hier}$. Another relevant difference is that in the \texttt{prioritized} $\xi$ method, contrary to PER, the gradients are not scaled, similar to a standard demonstration buffer.

Important to note that the formulation of HiER is fundamentally different from~\cite{schaul_prioritized_2016,oh_self-imitation_2018,ferret_self-imitation_2020,andrychowicz_hindsight_2017,wang_boosting_2019,bujalance2023reward}, not only but most importantly because of the idea of the secondary experience replay.

\subsection{E2H-ISE}\label{sec:method_cl}

A key attribute of HiER is that it learns from relevant positive experiences, described in Section~\ref{sec:method_hier}. However, if these experiences are scarce in the first place, $\mathcal{B}_{hier}$ would be considerably limited or even empty. Thus, HiER could benefit from an \textit{easy2hard} data collection CL method by having access to more positive experiences.

E2H-ISE is a data collection CL method based on controlling the entropy of the initial state-goal distribution $\mathcal{H}(\mu_0)$ and with it, indirectly, the task difficulty. In general, $\mu_0$ is constrained to one point (zero entropy) and moved towards the uniform distribution on the possible initial space (max entropy). Even though certain E2H-ISE versions allow decreasing the entropy, in general, they move $\mu_0$ towards max entropy.

\begin{figure}[h!tb]
\centering
\includegraphics[angle=0,width=0.7\linewidth]{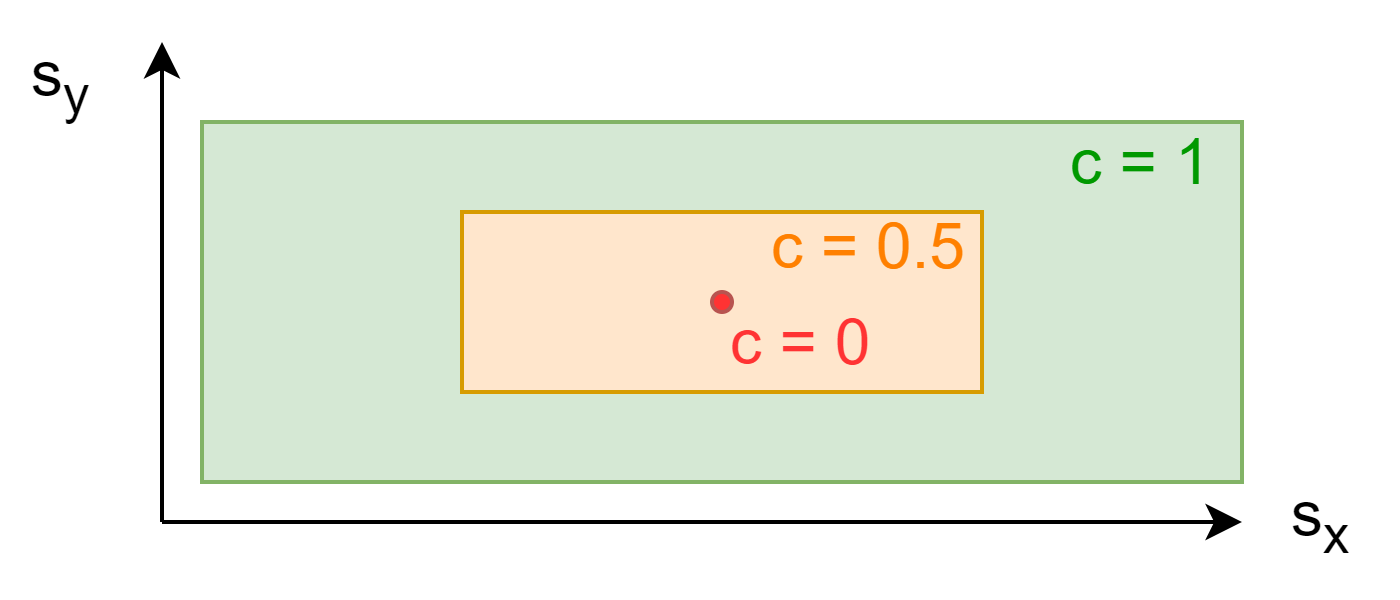}
\caption{Visualization of the effect of parameter $c$ on $\mu_0$ in a 2D case where state $s = [s_x,s_y]$. The initial state $s_0 = [s_{0,x},s_{0,y}]$ is sampled from the probability distribution $\mu_0(c)$.}
\label{fig:param_c}
\end{figure}

To formalize E2H-ISE, the parameter $c \in [0,1]$ is introduced as the scaling factor of the uniform $\mu_0$, assuming that the state space, including the goal space, is continuous and bounded.  The visualization of the scaling factor $c$ is depicted in Fig.~\ref{fig:param_c}. If $c = 1$ there is no scaling, while $c = 0$ means that $\mu_0$ is deterministic and returns only the center point of the space. To increase or decrease $\mathcal{H}(\mu_0)$, $c$ changes in time, thus we define $c_j$ for every $j$ where $j \in \mathbb{N}$ is the index of the episode. At the start of the training, $c$ is initialized and it is updated at the beginning of every training episode before $s_0$ is sampled from $\mu_0$.\footnote{For evaluation, the environment is always initialized according to the unchanged $\mu_0$.} The following versions are proposed for updating $c$: 

\begin{itemize}
    \item \texttt{predefined}: $c$ changes according to a predefined profile similar as in the case of $\lambda$ \texttt{predefined} (see Section~\ref{sec:method_hier}). In this paper, only the linear profile with saturation was considered.\footnote{We have experimented with a 2-stage version where $\mu_0$ and $\mu_G$ (initial goal distribution) were separated.}
    
    \item \texttt{self-paced}: $c$ is updated according to:
    \begin{equation}
    c_j =
    \begin{cases}
       \min(1,\ c_{j-1} + \delta) , & \text{if}\ P_{train,w} > \Psi_{high} \\
       \max(0,\ c_{j-1} - \delta) , & \text{if}\ P_{train,w} < \Psi_{low} \\
      c_{j-1}, & \text{otherwise}
    \end{cases}
  \end{equation}
  where $P_{train,w} \in \mathbb{R}$ is mean of last $w \in \mathbb{Z}^+$ (window size) training success rate, $\delta \in [0,1]$ is the step size, and $\Psi_{high} \in \mathbb{R}$ and $\Psi_{low} \in \mathbb{R}$ are threshold values.\footnote{If $\Psi_{low} = 0$, then $c$ can only increase.} After any update on $c$, $P_{train,w}$\footnote{The circular buffer storing the success rates.} is emptied, and the update on $c$ is restarted after $w$ episodes.
    \item \texttt{control}: $c$ is updated according to: 
    \begin{equation}
    c_j =
    \begin{cases}
       \min(1,\ c_{t-j} + \delta) , & \text{if}\ P_{train,w} \geq \psi \\
       \max(0,\ c_{t-j} - \delta) , & \text{if}\ P_{train,w} < \psi
    \end{cases}
    \end{equation}
    where $\psi \in \mathbb{R}$ is the target. The algorithm attempts to move and keep $P_{train,w}$ at $\psi$. Updates are executed only if $j > w$.
    \item \texttt{control adaptive}: This method is similar to \texttt{control} but the target success rate $\psi$ is not fixed but computed from the mean evaluation success rate:
    \begin{equation}
        \psi_j = \min \left( \psi_{max},\ \Delta + \frac{1}{w} \sum_{i=1}^{w}R_{j-i}^{eval} \right) 
    \end{equation}
    where $\Delta \in [0,1]$ is a constant shift (as we want to target a better success rate than the current) and $\psi_{max} \in \mathbb{R}$ is the maximum value allowed for $\psi$.\footnote{Important to note that contrary to the training, in the evaluation, we sample from the unrestricted $\mu_0$ ($c = 1$), thus the eval success rate represents the real success rate of the agent. Consequently, $c$ can be set to keep the training to a success rate that is just (by $\Delta$) above the eval success rate.} Updates are executed only if $j > w$. 
\end{itemize}

Sampling from $\mu_0(c \neq 1)$ introduces bias to the states within the probability distribution of $\mu_0(c)$. This bias is reduced as $c$ increases. Furthermore, as the buffers are circular, once they reach their capacity, the old experiences are replaced with new ones. On the other hand, we conducted experiments on dynamically subtracting the center of $\mu_0(c)$ to counterbalance the sampling bias, e.g.: $\mu_0(c) = \mu_0(c_1) - \mu_0(c_2)$ where $c_1 > c_2$. However, they did not result in any improvement. Our experimental results, presented in Section~\ref{sec:exps}, show that accepting the bias and starting with $c$ close to zero is beneficial as HiER+ further improves the performance of HiER.

Important to note, our E2H-ISE formulation is inherently different from  ~\cite{florensa_reverse_2018, ivanovic_barc_2019, salimans_learning_2018} as our solution does not concentrate on goal difficulty but the entropy of $\mu_0$. In our case, the \textit{easy2hard} attribute derives from the magnitude of the entropy and not from the goal difficulty. It is also disparate from~\cite{pong_skew-fit_2020} as their solution maximizes the entropy of goal-conditioned visited states $\mathcal{H}(\mathcal{S}|\mathcal{G})$ and not $\mathcal{H}(\mu_0)$. Nevertheless, the E2H-ISE method is only an example of data collection CL methods that can be utilized in HiER+. It is proposed in this paper, as it is significantly easier to implement than the presented, more sophisticated, state-of-the-art methods, while it is universal and requires minimal prior knowledge. Thus, the full potential of HiER+ can be presented conveniently with the E2H-ISE method. Comparing different data collection CL methods in HiER+ is out of the scope of this work.

\subsection{HiER+}\label{sec:method_HiER+}

In this section, HiER+ is presented which is an enhancement of HiER with an arbitrary data collection CL method. Even though in this work, we present HiER+ with E2H-ISE, it is important to note that the fundamental architecture of HiER+ would remain consistent when paired with alternative data collection CL approaches. It can be added to any off-policy RL algorithm with or without HER and PER, as depicted in Fig.~\ref{fig:HiER+} and presented in Algorithm~\ref{alg:HiER+}. Having initialized the variables and the environment (Lines 1-6), the training loop starts. After collecting an episode, its transitions are stored in $\mathcal{B}_{ser}$, and if HER is active then virtual experiences are added as well (Lines 13-14).\footnote{$\mathcal{B}_{ser}$ and $\mathcal{B}_{hier}$ are circular buffers, thus once they are full, the new transitions are replacing the old ones.} Then the $\lambda$ parameter of HiER is updated and if the given condition is met, the episode is stored in $\mathcal{B}_{hier}$ as well (Lines 15-18). In the next steps, the $c$ parameter of E2H-ISE is updated and the environment is reinitialized (Line 19-21), thus the agent can start collecting the next episode. At a given frequency, the weights of the models are updated (Line 23-31). The batches of $\mathcal{D}_{ser}$ and $\mathcal{D}_{hier}$ are sampled and combined (Lines 24-26). After the weight update (Line 27), if PER is active, the priorities in $\mathcal{B}_{ser}$ are updated (Line 28). Finally, the $\xi$ parameter and with it the batch size of HiER is updated (Lines 29-30).

\begin{algorithm}
\caption{HiER+}\label{alg:HiER+}
\begin{algorithmic}[1]

\State Initialize $c \gets 0,\ \lambda,\ \xi,\ n,\ \theta,\ \phi$
\State $n_{hier} \gets \xi \cdot n$ \Comment{n: batch size}
\State Initialize $\mathcal{B}_{ser} \gets \emptyset$
\State Initialize $\mathcal{B}_{hier} \gets \emptyset$
\State Initialize episode buffer $\mathcal{E} \gets \emptyset$
\State $s \gets \mu_0(c_0)$ \Comment{Init env}
    \While{Convergence}
    
        \State $a \gets \pi_{\theta}(s)$ \Comment{Collecting data}
        \State $s_2, r, d \gets$ Env.step($a$)
        \State $\mathcal{E} \gets \mathcal{E} \cup (s,a,s_2,r,d)$
        \State $s \gets s_2$
       
        \If{Episode ends}
            \State $ \mathcal{B}_{ser} \gets \mathcal{B}_{ser} \cup \mathcal{E}$ \Comment{Store transitions of $\mathcal{E}$}
            \State $ \mathcal{B}_{ser} \gets \mathcal{B}_{ser} \cup \mathcal{E}_{virtual}$ \Comment{HER (optional)}
            \State Update $\lambda_j$ \Comment{HiER: Section~\ref{sec:method_hier}}
            \If{$\lambda_j < \sum_{r_i \in \mathcal{E}}r_i$}
                 \State $ \mathcal{B}_{hier} \gets \mathcal{B}_{hier} \cup \mathcal{E}$
            \EndIf
            \State Update $c_j$ \Comment{E2H-ISE: Section~\ref{sec:method_cl}}
            \State $\mathcal{E} \gets \emptyset$
            \State $s \gets \mu_0(c_j)$      
        \EndIf
    
        \If{Weight update}
            \State $\mathcal{D}_{ser} \gets $ select $(n - n_{hier})$ sample from $\mathcal{B}_{ser}$
            \State $\mathcal{D}_{hier} \gets $ select $n_{hier}$ sample from $\mathcal{B}_{hier}$
            \State $\mathcal{D} \gets \mathcal{D}_{ser} + \mathcal{D}_{hier}$
            \State Update weights $\theta,\ \phi$ based on $\mathcal{D}$ 
            \State Update priorities in $\mathcal{B}_{ser}$ \Comment{PER (optional)}
            \State Update $\xi_k$ \Comment{HiER: Section~\ref{sec:method_hier}}
            \State $n_{hier} \gets \xi_k \cdot n$
        \EndIf  
    
        \If{Evaluation}
            \State Evaluate agent with $\mu_0(c=1)$ \Comment{Standard init}
        \EndIf
    \EndWhile
\end{algorithmic}
\end{algorithm}

\section{Results}\label{sec:exps}

Our contributions were validated on 8 tasks of three robotic benchmarks. The tasks are the push, slide, and pick-and-place tasks of the Panda-Gym~\cite{gallouedec_panda-gym_2021} and the Gymnasium-Robotics Fetch benchmarks~\cite{FetchEnv}, and two mazes, depicted on Fig.~\ref{fig:result_maze_tasks} of the Gymnasium-Robotics PointMaze environment~\cite{Maze}. The Panda-Gym Environment is based on the PyBullet~\cite{coumans2016pybullet} physics engine while the Gymnasium-Robotics Fetch and PointMaze environments are based on MuJoCo~\cite{mujoco}.

It is important to note, that for all tasks, the state and action spaces are continuous, and the reward function is sparse without any reward shaping.  Furthermore, the agent is devoid of access to any form of demonstration. These constraints, significantly exacerbate the difficulty of exploration.

The naming convention of the algorithms is the following: Algorithm~[Components]. The algorithm can be either Baseline, HiER, or HiER+, and the options for the components are HER and PER\footnote{With the exception of Fig.~\ref{fig:e2h_agg_metrics}.}. Thus, Baseline~[HER~\&~PER] means that the base (SAC, TD3, or DDPG) RL algorithm was applied with HER and PER. On the other hand, HiER~[HER] means that the base RL algorithm was applied with our HiER method and HER but without PER. HiER+ is HiER with E2H-ISE.

First and foremost, we present our evaluation protocol in Section~\ref{sec:result_reproduc} which is essential for result reproducibility. Then, the aggregate performance (across all tasks) of HiER is shown compared to their corresponding baselines in Section~\ref{sec:agg}. Subsequently, HiER and HiER+ (with E2H-ISE) are thoroughly evaluated on the push, slide, and pick-and-place tasks of the Panda-Gym robotic benchmark in Section~\ref{sec:panda-gym}. Furthermore, HiER is evaluated on the push, slide, and pick-and-place tasks and two mazes, depicted on Fig.~\ref{fig:result_maze_tasks} of the Gymnasium-Robotics Fetch and PointMaze benchmarks in Section~\ref{sec:exps_fetch} and Section~\ref{sec:exps_maze}. Then, the qualitative results of all tasks are evaluated in Section~\ref{sec:qualitative}. Additionally, the comparisons of the different $\xi$, $\lambda$, and $c$ methods are presented in Section~\ref{sec:exps_hier} and~\ref{sec:exps_e2h-ise}. Finally, our method is validated with DDPG and TD3 in Section~\ref{sec:exps_agents}.

For our experiments, the SAC RL algorithm was chosen, except in Section~\ref{sec:exps_agents}. The standard hyperparameters are set as default in \cite{spinning_up} except the discount factor $\gamma = 0.95$ as in \cite{gallouedec_panda-gym_2021}, and the SAC entropy maximization term $\alpha = 0.1$. The buffer size of $\mathcal{B}_{hier}$ was set to $10^6$.

In all the experiments with the exception of Section~\ref{sec:exps_hier} and~\ref{sec:exps_e2h-ise}, HiER was applied with the \texttt{predefined} $\lambda$ method and with the \texttt{prioritized} version of $\xi$ when PER was active and with the \texttt{fix} version with $\xi = 0.5$ otherwise. Furthermore, in HiER+, the E2H-ISE method was employed with the \texttt{self\--paced} option. The aforementioned settings were selected according to our comparison presented in Section~\ref{sec:exps_hier} and~\ref{sec:exps_e2h-ise}.

\subsection{Evaluation protocol}\label{sec:result_reproduc}

For results reproducibility, it is important to disclose the evaluation protocol. Each algorithm (configuration) and task pair is trained in 10 independent runs with different random seeds. For every run, at a specified frequency, the evaluation performance of the model is measured, presented at Line 32-34 of Algorithm~\ref{alg:HiER+}. The two most relevant performance metrics are the evaluation success rate and the evaluation accumulated reward, henceforth success rate and reward. In this paper, the performance is measured 50 times during a single training, and each time, the evaluation score is computed by taking the mean of 100 episodes. At the end of the training, all evaluation data is saved and stored. For the evaluation presented in this paper, in the case of success rates, the best scores of each run were the base datapoints\footnote{For calculating the mean, median, IQM, and OG scores.}, meaning for each algorithm (configuration) and task pair there are 10 datapoints, one for every run. This evaluation protocol follows~\cite{agarwal_optimistic_2020, badia_agent57_2020, mnih_human-level_2015} and the idea is similar to the method of early stopping.

In the following sections, the primary basis of evaluation is the success rate which was chosen for the following reasons:
\begin{itemize}
    \item Our main objective is to solve the tasks with the highest success rate. As we focus on sparse reward scenarios with Eq.~(\ref{eq:r}), the only additional information in the reward score is how fast the agent solved the task which is less relevant in our case.
    \item The success rate is an already normalized scale between zero and one. Reward scores of Eq.~(\ref{eq:r}) with different time horizons are significantly disparate.
    \item The reward value depends on the reward function itself. The same task can be executed with a different reward function, whose results are not comparable.
    \item The success rate could be seen as a specific reward function giving zero reward to every non-goal state, and one for every goal state.
\end{itemize}
Nevertheless, we report our reward scores, for the aggregated results, presented in Tab.~\ref{tab:results_agg_alltasks}, and for the results of HiER and HiER+ on the Panda-Gym environment, displayed in Tab.~\ref{tab:results_sota_reward}. In the cases of reward scores, instead of the best, the last values of each run were utilized. Our aim is to show that our methods outperform the state-of-the-art not only in the chosen evaluation protocol but in other protocols as well.

In general, we present our results with the mean, median, interquartile mean (IQM), and optimality gap (OG) metrics. For the former three, higher values are better, while for OG, the lower score is better. In the case of the success rate, the desired target is 1.0 which is the maximum achievable score\footnote{As the desired target is 1.0 which is in itself the highest possible number, these results are redundant as the mean is also presented. Nevertheless, to facilitate comparison, we preferred to keep them in the graphs.}. For displaying the amount of uncertainty, in the graphs, 95\% confidence intervals (CIs) were applied.

For plotting the figures of aggregated results, the performance profiles, and the probability improvements, the \textit{rliable}\cite{agarwal_deep_2021} library was utilized. Having 10 runs was sufficient, thus we present our results without task bootstrapping (as default in \textit{rliable}).

\subsection{Aggregated results across all tasks}\label{sec:agg}

Prior to showing the experimental results on each of the three robotic benchmarks, this section provides a summary of the aggregated results across all tasks, focusing on HiER and HiER~[HER].

Our experimental results are presented in Tab.~\ref{tab:results_agg_alltasks} and Fig.~\ref{fig:all_agg_metrics}. The results indicate that both HiER versions outperform their corresponding baseline, and HiER~[HER] yields the best performance in all metrics. In terms of point estimates, while Baseline~[HER] yields 0.56 and -43.7 IQM success rate and IQM reward, HiER~[HER] achieves 0.83 and -32.48 scores which are increments of 0.27 and 11.22, respectively. Moreover, regarding the uncertainty, both HiER and HiER~[HER] are superior to their corresponding baselines as the confidence intervals do not overlap.

Additionally, the performance profile graph, presented in Fig.~\ref{fig:all_performance_profile}, displays the run-score and the average-score distributions of the aforementioned algorithms. It shows that both HiER and HiER~[HER] have stochastic dominance over their baselines.

Finally, Fig.~\ref{fig:all_probability_hier} shows that both HiER and HiER~[HER] outperform their baselines with 0.85 and 0.88 probability\footnote{Important to note, that these probabilities could be significantly higher if the easy tasks were removed.}. Additionally, HiER~[HER] surpasses HiER with a probability of 0.76.

%\afterpage{\clearpage} % Ensure the next float is on a new page

\aggresults

\subsection{Panda-Gym}\label{sec:panda-gym}

Having presented the aggregated results on HiER, we present our results on the Panda-Gym robotic benchmark with more details and task-specific results. Additionally, we demonstrate how HiER+ with E2H-ISE can further improve the performance of HiER. From the Panda-Gym robotic benchmark, three robotic manipulation tasks were considered: 
\begin{itemize}
    \item \texttt{PandaPush-v3}: A block needs to be pushed to a target. Both the block starting position and the target position are within the reach of the robot.
    \item \texttt{PandaSlide-v3}: A puck needs to be slid to a target position outside of the reach of the robot.
    \item \texttt{PandaPickAndPlace-v3}: A block needs to be moved to a target that is oftentimes in the air thus the robot needs to grasp the block.
\end{itemize}

The starting position of the block (or the puck) and the goal position are sampled from the corresponding distributions. The action space is composed of incremental actions on the tool center point in $x$, $y$, and $z$ axes. Furthermore, in the case of the \texttt{Panda\-Pick\-And\-Place\--v3} task, the action space expanded with a continuous gripper control action. The reward function is sparse, as described in Eq.~(\ref{eq:r}). The tasks are depicted in Fig.~\ref{fig:result_panda_tasks}. For further details, we refer the reader to~\cite{gallouedec_panda-gym_2021}.

The aggregated results are presented in Fig.~\ref{fig:panda_agg_metrics}, while the performance profiles of the algorithms are demonstrated in the left side of Fig.~\ref{fig:panda_performance_profile_and_prob}. Our experimental results show that HiER (blue) and both versions of HiER+ (purple and magenta) significantly outperform the baselines (gray), while E2H-ISE alone could only slightly improve the performance. Moreover, the right side of Fig.~\ref{fig:panda_performance_profile_and_prob} shows at least a 0.99 average probability of improvement for our methods compared to the baselines.

Regarding the specific tasks, the learning curves of the selected configurations are depicted in Fig.~\ref{fig:result_panda_tasks}. For all cases, HiER and HiER+ significantly outperform the baselines. Moreover, Tab.~\ref{tab:results_sota_summary} presents a simplified summary of the performance of the algorithms on the specific tasks. Our results show that HiER~[HER] enhances its baseline by an increment of 0.03, 0.44, and 0.12 IQM score on the \texttt{Panda\-Push\--v3}, \texttt{Panda\-Slide\--v3}, and \texttt{Panda\-Pick\-And\-Place\--v3} tasks. Nevertheless, HiER+~[HER] further improves the performance, achieving 1.0, 0.82, and 0.71 IQM scores. Tab.~\ref{tab:results_sota} and Tab.~\ref{tab:results_sota_reward} display the results of all configurations based on their success rates and rewards.

\pandafigs

\pandatabs

\subsection{Gymnasium-Robotics Fetch}\label{sec:exps_fetch}

In this section, HiER is evaluated on the \texttt{Fetch\-Push\--v2}, \texttt{Fetch\-Slide\--v2}, and \texttt{Fetch\-Pick\-And\-Place\--v2} tasks of the MuJoCo-based Gymnasium-Robotics Fetch environment. 

Even though the tasks are similar to the Panda-Gym robotic benchmark, the robot configuration, the observation space, and the environment dynamic (different simulator) are disparate. Our goal with these experiments is to demonstrate that HiER does not uniquely work for the Panda-Gym robotic benchmark. The tasks are depicted in Fig.~\ref{fig:result_fetch_tasks}. For more details, we refer the reader to~\cite{FetchEnv}.

In this section, HiER and HiER~[HER] are compared with their corresponding baselines. Our experiment results are presented in Tab.~\ref{tab:results_fetch} and depicted in Fig.~\ref{fig:result_fetch_tasks} and Fig.~\ref{fig:fetch_agg_metrics}. In all cases, the HiER versions outperform their corresponding baselines. Regarding the \texttt{Fetch\-Push\--v2} task, HiER~[HER] improves the IQM score of the Baseline~[HER] method by 0.06 (increasing from 0.92 to 0.98). In the case of the \texttt{Fetch\-Slide\--v2} task, HiER achieves the best result with a 0.56 IQM score, yielding a 0.54 increase compared to its baseline with 0.02. Interestingly, adding HER worsens the performance. Nevertheless, HiER~[HER] still outperforms Baseline~[HER]. Finally, for the \texttt{Fetch\-Pick\-And\-Place\--v2} task, HiER~[HER] achieves a 0.73 IQM score. Compared to the Baseline~[HER] method with 0.24, it yields a 0.49 improvement. Interesting to note that for the latter two tasks, both HiER versions outperform both baselines.

\fetchresults

\subsection{Gymnasium-Robotics PointMaze}\label{sec:exps_maze}

In this section, HiER is evaluated on the \texttt{Point\-Maze\-Wall\--v3} and \texttt{Point\-Maze\--S\--v3} tasks of the MuJoCo-based Gymnasium-Robotics PointMaze environment to show the universality of our approach in a fundamentally different problem.

In these tasks, a ball, placed in a maze, needs to move from the start position to the goal position in a continuous state and action space. The start and the target positions are generated randomly with some constraints. For more details, we refer the reader to~\cite{Maze}. 

In our experiments, two different maze layouts were considered as depicted in Fig.~\ref{fig:result_maze_tasks}. The reward function is changed to Eq.~(\ref{eq:r}). As the tasks take longer to execute, the horizon is 500 timestep which is tenfold compared to the robotic manipulation tasks. Thus, for these experiments, the discount factor $\gamma$ was set to one\footnote{Not having a discount on future reward does not pose a problem as the reward function is formulated with -1 reward in every timestep, described in Eq.~(\ref{eq:r}). Thus, the agent aims to solve the task as fast as possible.}. 

The results of our experiments are presented in Tab.~\ref{tab:results_maze} and depicted in Fig.~\ref{fig:result_maze_graphs}. In the case of the \texttt{PointMaze\--Wall\--v3} task, the results are quite close to the optimal 1.0 success rate, thus there is no significant difference, even though HiER still performs equally or better than the baselines depending on the metrics and the configurations. Regarding the more challenging \texttt{PointMaze\--S\--v3} task, HiER~[HER] outperforms Baseline~[HER] by 0.2 IQM score, rising from 0.69 to 0.89.

\mazeresults

\subsection{Qualitative evaluation}\label{sec:qualitative}

In this section, the qualitative evaluation of the aforementioned tasks is presented. We refer the reader to the project site\footnote{\href{http://www.danielhorvath.eu/hier/\#bookmark-qualitative-eval}{http://www.danielhorvath.eu/hier/\#bookmark-qualitative-eval}} to watch our results compared with the baselines.

Regarding the Panda-Gym and the Gymnasium-Robotics Fetch environment, on many occasions, the baseline appears to be disoriented and incapable of completing the task. It appears that, during the training process, the agent became entrapped in a local minimum as a result of the challenging exploration problem caused by the continuous state and action space, the sparse reward, and the lack of demonstrations. This phenomenon is significantly less frequent in the case of HiER and HiER+ which solve the tasks with a considerably higher success rate, in correlation with the presented quantitative evaluation.

In the case of the Gymnasium-Robotics PointMaze environment, the qualitative evaluation does not show relevant differences. The primary reason is that while the mean, median, and IQM success rate scores are considerably higher in the case of HiER~[HER], both HiER~[HER] and Baseline~[HER] managed to obtain a perfect success rate of 100\% at least once in the PointMaze environment (see Tab.~\ref{tab:results_maze}).

\subsection{Other}\label{sec:exps_other}

In this section, the different $\lambda$, $\xi$, and $c$ methods are presented in Section~\ref{sec:exps_hier} and~\ref{sec:exps_e2h-ise}. Additionally, our method is validated with DDPG and TD3 in Section~\ref{sec:exps_agents}. All experiments were conducted on the Panda-Gym benchmark.
 
\subsubsection{HiER $\lambda$ and $\xi$ methods}\label{sec:exps_hier}

The comparison of the different HiER $\lambda$ methods are depicted in Fig.~\ref{fig:results_hier_modes}~(a) and~(b) and Fig.~\ref{fig:hier_lambda_agg_metrics}. The experiments were executed without HER, PER, and E2H-ISE. In these settings, the \texttt{predefined} $\lambda$ method outperforms the other variants, although its CI overlaps the CI of the \texttt{fix} $\lambda$ method. The $\lambda$ profiles are presented in Fig.~\ref{fig:results_hier_modes}~(b).

The impact of HiER $\xi$ method is shown in Fig.~\ref{fig:results_hier_modes}~(c) and Fig.~\ref{fig:hier_xi_agg_metrics}. The experiments were executed with HER and E2H-ISE but without PER. In these settings, the \texttt{fix} $\xi = 0.25$, $\xi = 0.5$, and the \texttt{prioritized} $\xi$ method appear to be the best versions in this order, although their CIs overlap\footnote{In other settings, we found $\xi = 0.5$ slightly better than the others.}. Important to note, that when PER is active, it scales the gradient proportionally to the probability of the samples, thus \texttt{prioritized} $\xi$ mode is recommended to counterbalance this effect.

\subsubsection{E2H-ISE versions}\label{sec:exps_e2h-ise}

The different E2H-ISE $c$ methods are presented in Tab.~\ref{tab:e2h} and displayed in Fig.~\ref{fig:e2h_agg_metrics}. The experiments were executed without PER. The ranking of E2H-ISE versions is relatively sensible for the applied methods (HER and HiER). Without HiER, there is no significant difference between the $c$ methods. With HiER but without HER the \texttt{control} and the \texttt{control} \texttt{adaptive} $c$ methods yield the highest performance, although their CIs overlap with the other versions. With HiER and HER, the \texttt{control} \texttt{adaptive} and \texttt{self-paced} $c$ methods achieve the best performance. Nevertheless, further optimization, or possibly another version of E2H-ISE could improve the performance.

\subsubsection{TD3 and DDPG}\label{sec:exps_agents}

To validate our methods not only with SAC, Fig.~\ref{fig:td3_ddpg_agg_metrics} and Tab.~\ref{tab:results_sota_td3_ddpg} show our results in the case of DDPG and TD3. In both cases, HiER+ improved the results of the baseline. In the case of TD3 (blue), the improvement is more significant as the CIs do not overlap. In the case of DDPG (magenta), although there is a considerable improvement, the CIs overlap.

\otherresults

\section{Conclusion}\label{sec:conclusion}

In this work, we introduced a novel technique called the highlight experience replay (HiER) to facilitate the training of off-policy reinforcement learning agents in a robotic, sparse-reward environment with continuous state and action spaces. Furthermore, the agent is devoid of access to any form of demonstration. These constraints, significantly exacerbate the difficulty of exploration.

In our method, a secondary replay buffer is created to store the most relevant experiences based on some criteria. At training, the transitions are sampled from both the standard experience replay buffer and the highlight experience replay buffer. Similarly to the hindsight experience replay (HER) and prioritized experience replay (PER), HiER can be added to any off-policy reinforcement learning algorithm. Following~\cite{portelas_automatic_2020}, HiER is classified as a data exploitation (or implicit) curriculum learning method.

To demonstrate the universality of HiER, it was validated on 8 tasks of three different robotics benchmarks~\cite{gallouedec_panda-gym_2021,FetchEnv,Maze} based on two different simulators~\cite{coumans2016pybullet,mujoco}. On one hand, among the 8 tasks, 3-3 were the same in principle (push, slide, and pick-and-place) but the robot configurations, the state spaces, and the dynamics of the environments were disparate. On the other hand, the last 2 tasks were fundamentally different as a ball needed to find a target in different mazes. 

In all of the experiments, HiER significantly improved the state-of-the-art methods. Our experimental results show that HiER is especially beneficial in hard-to-solve tasks such as \texttt{Panda\-Slide\--v3}, \texttt{Fetch\-Pick\-And\-Place\--v2}, or  \texttt{Point\-Maze\--S\--v3}.

HiER collects and stores positive experiences to improve the training process. With HiER+, we showed how HiER can benefit from a traditional, data collection curriculum learning method as well. Lack of general and easy-to-implement solutions, we proposed E2H-ISE, an \textit{easy2hard} data collection CL method that requires minimal prior knowledge and controls the entropy of the initial state-goal distribution $\mathcal{H}(\mu_0)$ which indirectly controls the task difficulty. Nevertheless, applying more sophisticated CL methods in place of E2H-ISE might be beneficial in future research.

HiER+ was validated on the \texttt{Panda\-Push\--v3}, \texttt{Panda\-Slide\--v3}, and \texttt{Panda\-Pick\-And\-Place\--v3} tasks of the Panda-Gym~\cite{gallouedec_panda-gym_2021} robotic benchmark. Our results show that HiER+ could further improve the performance of HiER.

Furthermore, we presented our experiments on the different $\lambda$, $\xi$, and $c$ methods of HiER and E2H-ISE. On one hand, we found that in the case of HiER $\lambda$, the \texttt{predefined} version was superior. On the other hand, the rankings of the $\xi$ and $c$  methods are more unambiguous and depend on the applied configuration. We also showed that HiER+ improves the baselines not only with SAC but with TD3 and DDPG as well.

Additionally, the qualitative analysis revealed that HiER and HiER+ showed a reduced tendency to be trapped in local minima compared to the vanilla baseline methods.

For future work, we will investigate other possible HiER versions. Moreover, we are interested in how HiER+ could facilitate sim2sim and sim2real knowledge transfer.

\clearpage

\bibliographystyle{IEEEtran}
\bibliography{hdlib}

\begin{IEEEbiography}[{\includegraphics[width=1in,height=1.25in,clip,keepaspectratio]{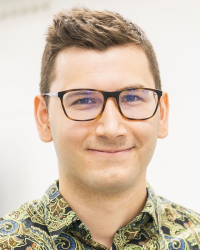}}]{Dániel Horváth (IEEE Member)} received his M.Sc.~degree with highest honours in mechatronics at the Budapest University of Technology and Economics, Hungary, in 2019. As part of his master's studies, he spent one semester each at the Technical University of Denmark in Copenhagen and at the Otto von Guericke University in Magdeburg, Germany, in 2018. 

He is pursuing his Ph.D.~at the Eötvös Loránd University, Budapest, Hungary in computer science in collaboration with the Institute for Computer Science and Control, Budapest Hungary, and, as a Campus France scholar, with  MINES Paris-PSL, in Paris, France under the supervision of Zoltán Istenes, Gábor Erdős, and Fabien Moutarde. His main research areas are reinforcement learning, curriculum learning, transfer learning, computer vision, and robotics. 

\end{IEEEbiography}

\begin{IEEEbiography}[{\includegraphics[width=1in,height=1.25in,clip,keepaspectratio]{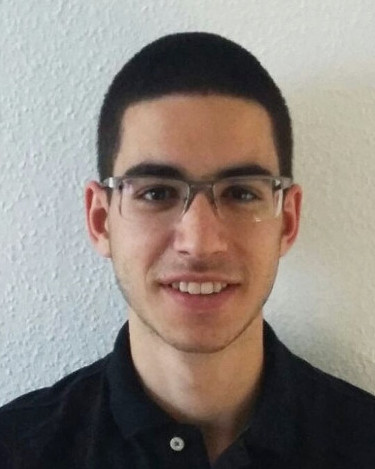}}]{Jesús Bujalance Martín} received his M.Sc. degree in mathematics and computer vision at IP Paris - Télécom ParisTech and Paris-Saclay University - ENS Cachan (M.Sc. MVA), in 2019.  

As part of his master’s studies, he spent one semester at Shanghai Jiao Tong University in 2018. He is pursuing his Ph.D. at MINES Paris-PSL under the supervision of Fabien Moutarde. His main research areas are reinforcement learning, computer vision, and robotics.
\end{IEEEbiography}

\begin{IEEEbiography}[{\includegraphics[width=1in,height=1.25in,clip,keepaspectratio]{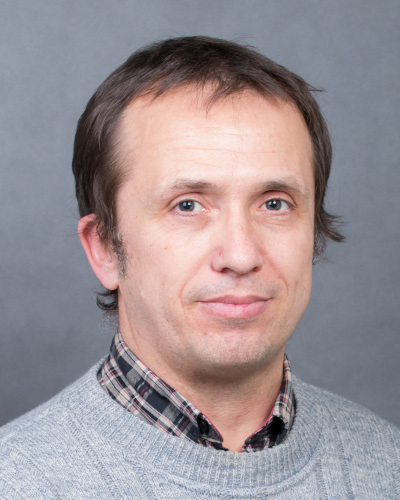}}]{Gábor Erdos} received his M.Sc.~degree in mechanical engineering at the State University of New York at Buffalo in 1995 and his Ph.D.~degree at the Budapest University of Technology and Economics, Hungary, in 2000. He was a postdoctoral researcher at the École Polytechnique Fédérale de Lausanne, Switzerland until 2002. 

He joined the Institute for Computer Science and Control, Budapest, Hungary as a researcher in 2003 and since 2018, he is the deputy head of the Research Laboratory on Engineering and Management Intelligence. His main research fields are robotics, point-cloud and 3D modeling, digital twin models, multi-body kinematics, and virtual manufacturing.
\end{IEEEbiography}

\begin{IEEEbiography}[{\includegraphics[width=1in,height=1.25in,clip,keepaspectratio]{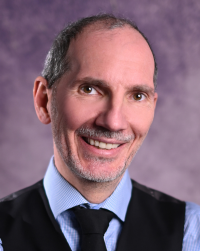}}]{Zoltán Istenes} received his Ph.D.~degree in Informatics in 1997 at the University of Nantes, France.

He is an associate professor at the Faculty of Informatics, Eötvös Loránd University (ELTE) in Budapest, Hungary. He established the ELTE Informatics Robotics Lab and currently directs the Erasmus Mundus Joint Master in Intelligent Field Robotic Systems (IFROS) in Hungary. At the European Institute of Innovation and Technology (EIT Digital), he manages the Budapest Doctoral Training Centre. His expertise encompasses a spectrum of fields including computer architectures, artificial intelligence, and robotics, with a recent focus on IoT, UAVs, and autonomous self-driving vehicles.
\end{IEEEbiography}

\begin{IEEEbiography}[{\includegraphics[width=1in,height=1.25in,clip,keepaspectratio]{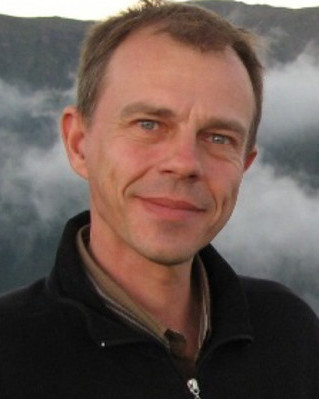}}]{Fabien Moutarde}, after graduating from Ecole Polytechnique, Paris, France in 1987, received his PhD degree in astrophysics in 1991 at Paris-Diderot University, Paris VII, France, and has obtained his "Habilitation to supervise PhDs" (HdR) in Engineering Sciences in 2013 at Pierre et Marie Curie University, Paris VI, France.

He joined MinesParis-PSL in Paris, France in 1996, where he is full professor since 2015, and is currently the director of the Center for Robotics. His research is centered on artificial intelligence and machine learning for robotics, particularly computer vision and reinforcement learning, with a focus on intelligent vehicles, as well as mobile or/and collaborative robots.
\end{IEEEbiography}
\EOD

\end{document}